# Harnessing The Power of Attention for Patch-Based Biomedical Image Classification

Gousia Habib[*1], Shaima Qureshi[2], Malik ishfaq[3]

1. *Postdoctoral Researcher Indian Institute of Technology New Delhi, Hauz Khas Delhi, 110016*
2. *Department of C.S.E., National Institute of Technology Srinagar 19006*
3. *Department of Mathematics University of Kashmir Srinagar 190006*

*Abstract:* *Biomedical image analysis is of paramount importance for the advancement of healthcare and medical research. Although conventional convolutional neural networks (CNNs) are frequently employed in this domain, facing limitations in capturing intricate spatial and temporal relationships at the pixel level due to their reliance on fixed-sized windows and immutable filter weights post-training. These constraints impede their ability to adapt to input fluctuations and comprehend extensive long-range contextual information. To overcome these challenges, a novel architecture based on self-attention mechanisms as an alternative to conventional CNNs.The proposed model utilizes attention-based mechanisms to surpass the limitations of CNNs. The key component of our strategy is the combination of non-overlapping (vanilla patching) and novel overlapped Shifted Patching Techniques (S.P.T.s), which enhances the model's capacity to capture local context and improves generalization. Additionally, we introduce the Lancoz5 interpolation technique, which adapts variable image sizes to higher resolutions, facilitating better analysis of high-resolution biomedical images. Our methods address critical challenges faced by attention-based vision models, including inductive bias, weight sharing, receptive field limitations, and efficient data handling. Experimental evidence shows the effectiveness of proposed model in generalizing to various biomedical imaging tasks. The attention-based model, combined with advanced data augmentation methodologies, exhibits robust modeling capabilities and superior performance compared to existing approaches. The integration of S.P.T.s significantly enhances the model's ability to capture local context, while the Lancoz5 interpolation technique ensures efficient handling of high-resolution images. This Study presents a novel approach for biomedical imaging, addressing key limitations of traditional CNNs. The implementation of attention-based models in biomedical image analysis can transform the field by providing more accurate and adaptable tools for medical research and healthcare.*
*.*

**Impact Statement:** The main aim of the research is to replace computationally complex convolution neural networks with strong modelling capacity attention-based models for biomedical image analysis. The study explores the direct implementation of attention-based models in biomedical image recognition without any dependence on any image-specific inductive biases offered by convolution. The research study includes Novel techniques, such as Lancoz5 interpolation and shifted patch tokenization, to introduce inductive bias and make these models adaptive to high-resolution images.

Social Impact: The more advanced A.I.A.I. model can serve the health sector in detecting and classifying different types of life-threatening diseases, including brain tumours, interstitial lung disease etc. The model can help in the early detection of infection with 100% accuracy. Early detection of these life-threatening diseases with incredible accuracy can save many precious human lives.

## I. INTRODUCTION

CNNs have long dominated the field of Computer Vision. They use filters to create feature maps highlighting the most relevant parts of the input image to generate simplified versions. A multilayer perceptron then uses the features to classify [1]. A recent revolution in this field was brought about by attention-based models, which through a mechanism of self-attention [2], achieved excellent results on a wide variety of tasks. This deep learning architecture is based primarily on a self-attention module and was developed mainly for sequence-to-sequence tasks (e.g., translating sentences between languages [3].

Contrary to CNNs, attention mechanisms dynamically learn and assign different levels of importance to other spatial locations within an image. By capturing long-range dependencies and intricate relationships between regions, models can focus on salient features while ignoring irrelevant information, thanks to self-attention, a prominent form of attention. This adaptability allows attention-based models to excel at tasks requiring global context comprehension, like image segmentation and object recognition, where the role of spatial relationships and semantic dependencies is crucial. Applying attention-based models to Computer Vision enables enhanced contextual understanding and adaptive feature weighting.

Deep learning research has recently achieved impressive results by adapting this architecture for computer vision tasks, such as image classification. The attention-based models applied to this domain are known (unsurprisingly) as Vision Transformers (ViTs). It works on the principle of self-attention, utilizing vectors extracted from the input image by splitting it into patches and projecting them into linear space. When given a sequence of items, self-attention estimates how relevant each item is to the rest (for instance, which words are likely to pair together in a sentence). As part of Transformers, the self-attention mechanism explicitly models interactions between entities of a sequence for structured predictions. A self-attention layer aggregates global information from the complete input sequence to update each component. As one of the most promising approaches in Computer Vision, attention-based architectures are achieving outstanding results. As an alternative to approaches conjugating attention-based components with CNN architectures, a model ViT utilizes the pure attention-based architecture directly to the sequence of images was proposed in [4]. Vision Transformer (ViT) divides the underlying image into patches, which are flattened, and then projected (linearly) into a fixed dimension. Next, each image patch is embedded with a position embedding indicating its location within the image. In ViT, the input embedding is processed simultaneously and identically, regardless of the sequence order. A standard method of utilizing sequential information is that an extra positional vector is appended to the inputs, hence the term "positional encoding"[5]. Various techniques are available for positional encoding. A simple example is given:

$$PE_{(ps,j)} = \begin{cases} \sin(ps \cdot w_k) & if\ j=2k \\ \cos(ps \cdot w_k) & if\ j=2k+1 \end{cases} \quad (1)$$

$$\text{where } w_k = \left(\frac{1}{1000}\right)^{\frac{2k}{l}},\ k=1,2,\ldots,\frac{l}{2} \quad (2)$$

*ps* and *l* are the position and length of the vector, respectively, and *j* is the index of each element within the vector. Furthermore, a 2D interpolation complements pre-trained positional encoding to keep the patches in the same order even when the input images have arbitrary positions. When compared with the most popular CNNs, ViT has achieved comparable or even superior results on multiple image recognition benchmarks (ImageNet [7] and CIFAR-100 [6]) by pretraining with a large-scale private dataset of Google (JFT-300M [8]). The ViT is equipped with two key elements: M.H.S.A. and F.F.N. Dong et al. [9] suggest that the Transformer can further mitigate the strong bias of M.H.S.A. by utilizing skip connections and F.F.N. in conjunction with the convolutional layer [10].

Introducing attention-based models allows for significant parallelization and quality improvement of translations using self-attention mechanisms and transformer architectures. In contrast to traditional recurrent neural networks (R.N.N.s) or convolutional models, attention-based models can process all positions in an input sequence simultaneously. An attention-based model generates each output element using self-attention to weigh the importance of different parts in the input sequence. It reduces the reliance on sequential processing by capturing long-range dependencies and relationships between words. Therefore, attention-based models can process multiple dishes simultaneously, resulting in significantly faster training and inference times than sequential models.

Furthermore, by incorporating multi-head self-attention into the transformer architecture, the model can simultaneously focus on different aspects of the input sequence to enhance translation quality. Depending on the position of the attention heads in the input, different linguistic patterns and dependencies will be captured. Due to this, the model can better grasp contextual nuances from the source language and capture relevant information. Attention-based models have revolutionized machine translation by accelerating training and inference by parallelizing and improving quality.

The entire paper is divided into fourteen different sections. Sections I and II give a detailed introduction to the self-attention mechanism and background of CNN and self-attention. Section III discusses the significant limitations of the C.N.N.S. and the motivation behind attention-based models for vision tasks. Section IV--XI provides extensive experiment details, novel methodologies used in the experimentation section, and results and discussion. Finally, the paper has done ablation studies in the XII section. Then the document also highlights the significant limitations of the study in the XIII section and suggests some future directions.

II. MOTIVATION

CNNs work on a fixed-sized window and struggle to capture spatial and temporal relations at the pixel level. In addition, CNN filter weights remain fixed after training, making it impossible for the operation to adapt dynamically to changes in input. In response to the shortcomings of CNNs, attention-based and Hybrid models were developed that combine the strengths of both models. It is no secret that Attention-based Transformer architectures are taking over the computer vision domain and becoming a popular choice in research and practice. Introducing these attention-based models enabled significant parallelization and optimization of translation quality. The shortcomings of CNN's forced me to replace the convolutions and shift towards attention-based models. These models have a strong modelling capacity due to more access to global information (a large receptive field) than CNN. Also, ViT skip connections have more potent effects on performance and representation similarity than ResNets skip references. For similar representations to be computed, the lower layers of ResNet had to be more complex than ViT. A larger pretraining dataset leads to significantly stronger intermediate terms in larger ViT models. The transformer network originated in machine translation and has the advantage of modelling long-range dependencies within a long sequence. The self-attention mechanism of transformers can adequately model global interactions between token embeddings. However, a locality mechanism for information exchange exists only within local regions. The locality is crucial for images since it pertains to structures like lines, edges, shapes, and even objects.

In light of this breakthrough, we need to answer specific questions.

- ✓ How are attention-based models solving these image-based tasks?
- ✓ Are they similar to convolutions, learning inductive biases from the ground truth?
- ✓ Do they develop new representations of tasks?

In the process of learning these representations, how does scale play a role? Do downstream tasks suffer as a result? The main Highlights of the work are given as:

This paper examines these questions, illuminating the critical representational differences between attention-based mechanisms and CNNs, and how these differences are derived. The article also throws light on adapting this attention mechanism to variable image size and resolution and how to bring locality into the attention-based vision models. Specifically, our significant contributions are:

- Explore the direct application of attention-based models to image recognition, where they do not introduce any image-specific inductive biases into the model architecture.

- Interpret an image as a sequence of patches and process it by a standard attention-based encoder.

- Preprocessing of Input medical images using different interpolation techniques and comparing with proposed novel lancoz5 interpolation to adapt the attention-based model to be more adaptable to variable size images with better resolution.

- We are introducing the locality using S.P.T. into the proposed attention-based model to generalize well on downstream tasks.
- We are verifying the generalization results on varying the number of patches and image sizes.

- Perform ablation studies by experimenting with different interpolation methods.

III. RELATED WORK

**CNNS:** A convolutional network (ConvNet) incorporates down sampling, shift invariance, and shared weights, making it the de-facto standard for computer vision tasks for images [11-21] and videos [22-34].

**CNNS and Self Attention**: Convolutional Neural Networks (CNNs) extract hierarchical features from visual data through localized convolutional operations and layer pooling. In object recognition and image classification tasks, CNNs are excellent at capturing spatial hierarchies and local patterns within images and videos. Alternatively, self-attention mechanisms provide a long-range contextual understanding by varying the weight of each position in a sequence based on context. With its ability to capture global relationships, self-attention is particularly valuable for capturing interaction and dependencies between input sequences, contributing significantly to machine translation and semantic segmentation tasks requiring an understanding of intricate relationships and context. The use of self-attention mechanisms in ConvNets has been studied to understand images [35-38], unsupervised object recognition and vision and language [39, 40]. There is evidence for the use of self-attention mechanisms to understand images and recognize objects [41 44], as well as in the areas of vision and language [45, 46].

**Vision Transformers:** Convolutional networks and self-attention operations have also been combined for video recognition and image understanding [47]. The current enthusiasm for using a standalone attention mechanism came into existence in the form of Transformers [48] to solve vision problems began with the Vision Transformer (ViT) [49] and the Detection Transformer [50]. In this paper, we extend [49] with a proposed model that enables variable resolution by down sampling with the Lancoz 5 interpolation technique for detecting and classifying biomedical images.

**Spatial Patch Tokenization**: A spatial patch tokenization technique is also employed for inducing inductive bias in the vision transformer models so that they can be generalized even to small datasets. Two novel techniques are presented to increase the locality inductive bias of ViT when training it on small datasets. The first aspect of S.P.T. involves embedding rich spatial information into visual tokens through a specific transformation. A second effect of L.S.A. is to induce ViT to attend locally using a softmax with learnable parameters. These two methods can be used independently and improve ViT performance [48-50].

**Data-Efficient Transformers (DeiT):** The DeiT [51] framework proposes a data-efficient model for training ViT. It is emerging to apply transformers to vision tasks such as image classification [52-53] object detection [54], semantic segmentation [55-56], 3D reconstruction [57], pose estimation [58], generative modelling [59], image retrieval [60], segmentation of medical images [61-63], point clouds [64], segmentation of video instances [65] dialogue of videos [68], and detection of video objects [66].

Several recent studies have reduced the quadratic attention complexity to optimize transformer performance for natural language processing applications, which complements our approach [67].

**Proposed Technique:** In contrast, our proposed patch-level attention-based model introduces locality via spatial patch tokenization., making it possible to work without convolutions and can be rescaled to large image sizes with better resolutions. Also, this paper provides a holistic view of how patch-level self-attention mechanism-based models are employed for computer vision tasks without being dependent on convolutions.

IV. PROPOSED METHODOLOGY

The generic frame of the proposed methodology can be best explained in Figure 1. The proposed model is typically made up of an attention-based encoder with layer-norm and M.L.P. head. We consider only the encoders as part of the network to perform image classification here. As a result, we will focus primarily on the operations performed in the encoder layers. The detailed description of each component of the proposed model is explained in the below sections:

**Algorithm for Proposed Methodology**

**Input:** BRAIN MRI Images

1. *Identifying Limitations of CNN.*
2. *We are introducing Self Attention as an alternative to CNN for capturing long-range contextual dependency.*
3. *Innovative techniques for injecting inductive bias into self-attention to make it behave like convolutions (Cut Mix, Lancoz 5 and S.P.T. tokenization scheme.*
4. *Perform extensive experiments to verify these techniques' impact on the proposed model's performance.*
5. *We are using attention-based models to reimagine convolution.*
6. *Using biomedical images without convolutions demonstrates the feasibility of classification.*
7. *Demonstrate the potency of attention-based vision models as effective modelling alternatives.*
8. *Summarize how an attention-based algorithm has replaced CNN limitations to conclude the algorithm.*

Output: The enhancement of vision analysis with attention-based models

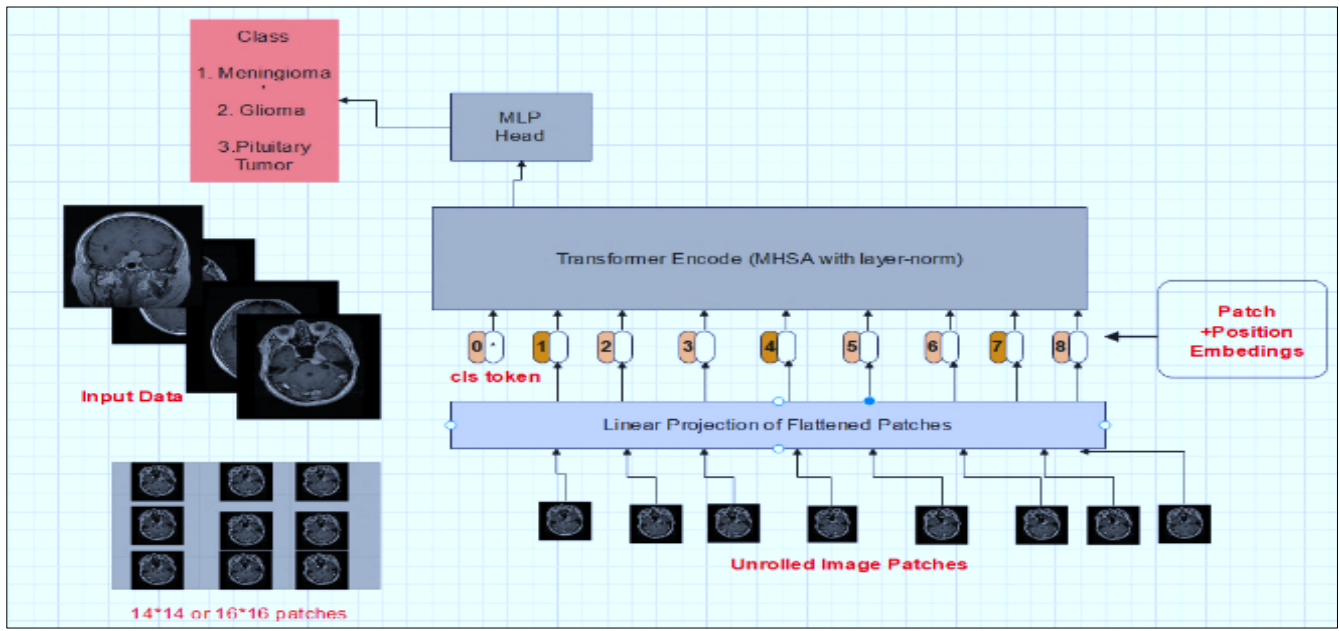

Figure.1 Framework of the patch-level attention mechanism

*a). Pacifying Input Image*

In image classification, ViT is a pure attention-based model that is directly applied to sequences of image patches. The generic framework of the ViT used in this research is shown in Figure 1. To process 2D images, the input image $\mathbf{x} \in \mathbf{R}^{H \times W \times C}$ is transformed into an array of flattened 2D patches $\mathbf{x}_p \in \mathbf{R}^{N \times (P^2 \cdot C)}$. $C$ represents the number of channels and $(H \times W)$ denotes the resolution of the original image. *(P, P)* represents the patch dimension of each image patch extracted from the input image. As a result, the effective sequence length for the model is $N = \frac{H \times W}{P^2}$. ViTs use constant latent vector size in all their layers, so the patch embeddings produced by trainable linear projection map each vectorized path to the model dimension d. An embedding patch sequence is embedded with a learnable embedding similar to the [class] token in B.E.R.T., an attention-based model used for N.L.P. tasks. This embedding's state serves as an image representation. The classification heads are the same size during the pretraining and fine-tuning tuning stages. A note on ViT is that it only uses the encoder of the standard Transformer, which precedes an M.L.P. head (except for the layer normalization). ViT is often trained on large datasets before fine-tuning for smaller downstream datasets.

Moreover, the self-attention part of the ViT captures only global dependencies between tokens. In other words, the attention block does not provide a mechanism for modelling local dependence between adjacent pixels. This would be interesting if the locality could be efficiently conveyed to these attention-based models. In this paper, we are employing two different patching techniques: Vanilla and S.P.T. The S.P.T. transforms an input image by spatially shifting it in several directions and then concatenates them with the input image. The next step is applying patch partitioning, just as in standard ViTs. Following patch flattening, layer normalization, and linear projection, three steps are sequentially performed to embed the tokens into visuals. More spatial information can be implanted through S.P.T. into visual tokens, increasing their locality inductive bias. Information from local and global sources is helpful when understanding how the contents of an image relate to each other. In addition to S.P.T., self-attention and both contribute to visual perception. There are two components. M.H.S.A. and layer-norm.

*b). Multi-Headed Self Attention (M.H.S.A.)*

ViT implements scaling of dot-product attention, similar to the general attention mechanism in the case of NLP-based models for each query, q, the scaled dot-product attention first computes a dot product with all the keys, k. This is followed by a softmax function which divides each result by the number of keys, $\sqrt{dk}$. Consequently, it obtains the weights used to scale the values, v. As a result, the computations performed by scaled dot-product attention can be applied efficiently to the entire pool of queries simultaneously. As inputs to the attention function, Q, K, and V are used as follows:

$$\text{Attention}(Q, K, V) = \text{Softmax}\left(\frac{QK^T}{\sqrt{d_k}}\right) \quad (3)$$

The multi-head attention mechanism linearly projects queries, keys, and values h times, using a different learned projection every time. To produce a final result, the single attention mechanism is applied to each of these h projections in parallel, producing h outputs, which are then concatenated and projected again. The generic idea can be best represented by Figure 2 [68]

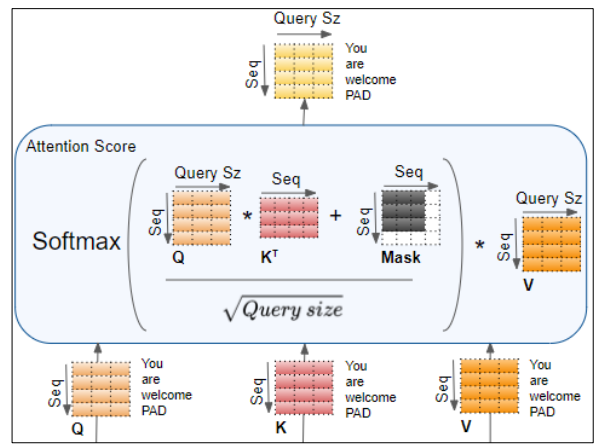

Figure 2. Self-Attention Mechanism

A multi-head attention model can attend to information from multiple representation subspaces from multiple positions simultaneously. In the case of a single attention head, averaging prevents this from happening. The general equation for the M.H.S.A. is given as follows:

$$\begin{aligned} \text{MHSA}(Q,K,V) &= \text{Concat}(head_1,\cdots,head_h)W^O \\ \text{where} \quad head &= \text{Attention}(QW_i^Q, KW_i^K, VW_i^V) \end{aligned} \quad (4)$$

Parameter matrices as $W_i^Q \in \mathbb{R}^{d_{model} \times d_k}$, $W_i^K \in \mathbb{R}^{d_{model} \times d_k}$ and $W_i^V \in \mathbb{R}^{d_{model} \times d_v}$ are the linear projections of Q, K and V with $W^O \in \mathbb{R}^{hd_v \times d_{model}}$ is the initial trainable matrix jointly trained with the model. The generic idea of the M.H.S.A. can be best mimicked by Figure 3

*c). Layer Norm (L.N.L.N.):*

It is added before each block to prevent new dependencies between training images. In this way, training time and overall performance are improved.
Other components of the proposed method are given [68]:

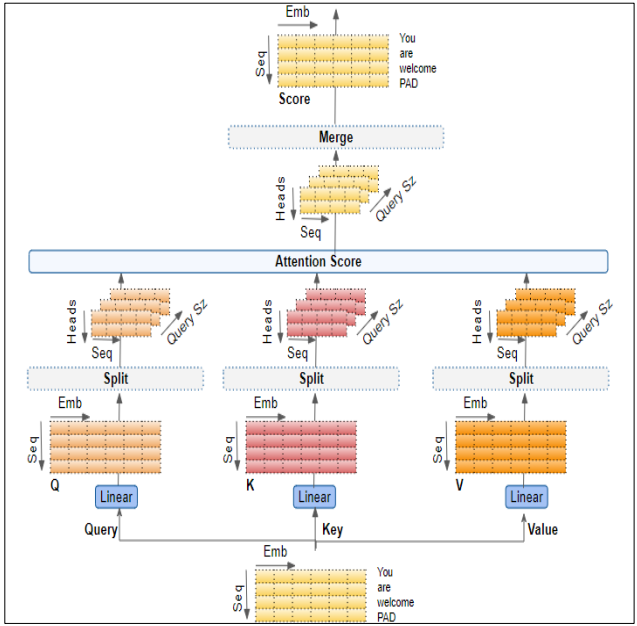

Figure 3: M.H.S.A. Mechanism

*d). Multilayer Perceptron (M.L.P.) Layer:*

This layer comprises two fully connected layers, each containing a Gaussian Error Linear Unit (G.E.L.U.).

*e). Embeddings and Softmax:*

The purpose of this section is to describe how Vanilla patching and S.P.T. can be used to embed/encode the patch embeddings.

*Vanilla 1-D positional embedding:* Considering the inputs as a sequence of patches in the raster order (default across all other experiments in this paper) and, another time, considering inputs as two-dimensional patches. Two sets of embeddings are learned here, each for one of the axes. The first embedding is for *X*, the second is for *Y*, and both have size *D*/2. To get the final positional embedding for the patch, we concatenate the X and Y embeddings based on the path coordinates in the input.

*f). S.P.T. position embeddings*

As a first step, each input image is spatially shifted by half the patch size in four diagonal directions: left-up, right-up, left-down, and right-down. As a convenience, in this paper, we refer to this shifting strategy as $T_{spatial}$, and in all experiments, we use $T_{spatial}$ as the S.P.T. The shifted features are cropped to match the input image and concatenated in the next step. The concatenated features are then divided into non-overlapping patches and are flattened. In the next step, layer normalization (L.N.L.N.) and linear projection are applied to obtain visual tokens. This process is described by Eq.5:

$$S_{pe}(\mathbf{x}) = \begin{cases} [\mathbf{x}_{cls}; S(\mathbf{x})] + \mathbf{E}_{pos} & \text{if } \mathbf{x}_{cls} \text{ exist} \\ S(\mathbf{x}) + \mathbf{E}_{pos} & \text{otherwise} \end{cases} \quad (5)$$

Where $\mathbf{x}_{cls} \in \mathbb{R}^{d_S}$ is a class token, and $\mathbf{E}_{pos} \in \mathbb{R}^{(N+1), \times d_S}$ represents learnable positional embedding. ds represent the transformer latent vector size. Ns denotes several image shifts shifted by Tspatial. Cls is the class token.

For positional encodings, both sines, as well as cosine functions at different frequencies are used:

$$PE_{(position, 2k)} = \sin\left(\frac{position}{10000^{\frac{2k}{d_{model}}}}\right)$$

$$PE_{(position, 2k+1)} = \cos\left(\frac{position}{10000^{\frac{2k}{d_{model}}}}\right) \quad (6)$$

Whereas position denotes the position, $k$ represents the dimension, and d represents the model depth.

*g). Softmax*

Using this function, a vector of numbers is converted into a vector of probabilities in which the probability of each value in the vector is proportional to its relative scale. In simpler terms, the logits of the network are converted to probability distribution scores.

V. EXPERIMENTAL SETUP

All the experiments are performed on standard Tesla GPU p100 with 32GB RAM using Google Colab supporting all the TensorFlow, Keras and Sckitlearn libraries. The proposed method is implemented on the medical image dataset given as:

*a). Dataset Description:*

This dataset contains 3064 T1-weighted contrast-enhanced images from 233 patients with three types of brain tumours: meningioma (708 slices), glioma (1426 slices), and pituitary tumour (930 slices). The entire description of the dataset is given below.
The following labels are found in the **data. Label file:** 1 for meningioma, 2 for glioma, and 3 for a pituitary tumour. **cjdata.** P.I.D.: patient identification number. **cjdata. Image**: the image data. In **data. Tumour border:** Discrete point coordinates are stored. **cjdata. tumorMask**: binary image containing 1s indicating tumour regi**ons. The** entire dataset is divided into three different sets Train (0.80), Test (0.10) and validation (0.10), followed by cut mix data augmentation technique.

*b). Cut-Mix data augmentation*

Data augmentation technique Cut Mix addresses the problem of information loss and inefficiency present in regional dropout strategies. Patches are randomly cut and pasted onto another image, and ground truth labels are mixed proportionally to the size of the patches. As an alternative to removing pixels and filling them with black pixels or Gaussian noise, a patch from another image can replace the removed regions. At the same time, ground truth labels are mixed proportionally to the number of pixels in the combined images. The more straightforward idea of Cut-Mix data augmentation depends upon the Cut-Mix function, which is given as follows:

> *Cut Mix Function*
> ✓ **The Cut Mix function takes two images and label pairs to perform the augmentation.**
> ✓ **The function takes a random sample from the Beta distribution and returns a bounding box using the get_box function.**
> ✓ **Afterwards, we crop and pad the second image (image2) in the final padded image at the exact location.**

As a result, we can generate the following virtual examples:
$$\tilde{\xi} = M\xi_i + (1-M)\xi_j$$
$$\tilde{\eta} = \lambda\eta_i + (1-\lambda)\eta_j \quad (7)$$

An array $M$ of binary masks (often squares) indicates the cut-out and fill-in regions for the two randomly drawn images. Just like mix-up $\lambda$ is removed from Beta $(\alpha, \alpha)$ distribution and $\lambda \in [0,1]$. When the images are randomly selected, bounding box coordinates are sampled so that both images indicate Cut-out and Fill-in regions.

$$r_\xi \sim U(0,W), r_w = W\sqrt{1-\lambda}$$
$$r_\eta \sim U(0,H), r_h = H\sqrt{1-\lambda} \quad (8)$$

$r_x, r_y$ Denotes the random samples drawn from a uniform distribution with an upper bound, as shown. Let us implement this augmentation strategy.

*c). Preprocessing*

In the preprocessing step, followed by data augmentation. Normalization and resizing of the images are done. Then interpolation step is applied to these resized images. This paper proposes the novel Lancoz 5 interpolation method, allowing large image sizes with better resolution. The detailed description of Lancoz 5 interpolation is given as follows:

*d). Lancoz 5 interpolation*

Lancoz5 maintains details and generates a few aliasing artefacts for geometric transformations that do not involve substantial down-sampling. Lanczos interpolation function of order n in one dimension is defined as follows:

$$\text{Lancoz5}(y; m > 0)$$

$$= \begin{cases} \text{sinc}(y) \cdot \text{sinc}\left(\dfrac{y}{m}\right) & \text{for } |y| \leq m \\ 0 & \text{otherwise} \end{cases} \quad (9)$$

A normalized sinc function is defined as:

$$\text{Since }(y) = \begin{cases} 1 & \text{for } y = 0 \\ \dfrac{\sin(\pi y)}{\pi y} & \text{otherwise} \end{cases} \quad (10)$$

To interpolate a two-dimensional image F using a Lanczos filter of mth order, the following algorithm is used:

$$f(x, y) = \frac{1}{w} \sum_{i=-n+1}^{n} \sum_{j=-n+1}^{n} f(\lfloor x \rfloor + i, \lfloor y \rfloor + j) \cdot L(i - x + \lfloor x \rfloor; n) \cdot L(j - y + \lfloor y \rfloor; n) \quad (11)$$

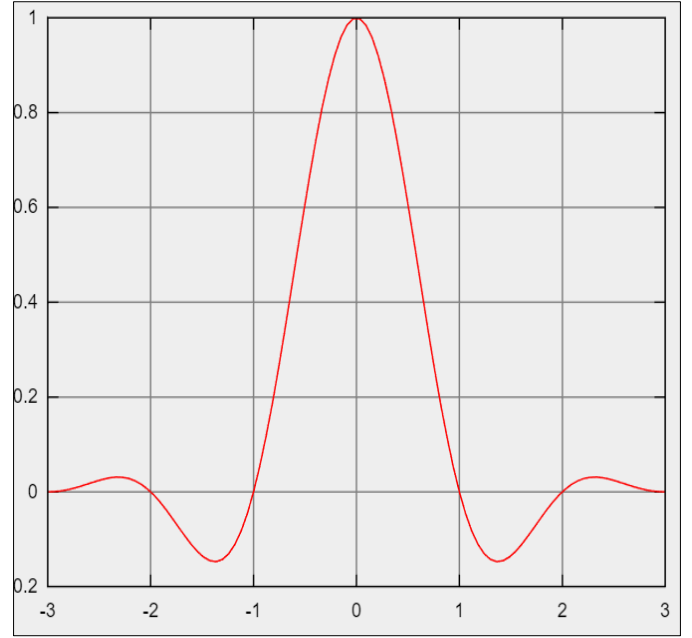

Figure 4. Lancoz interpolation of order 3.

Where $(x, y)$ represents the coordinates of the interpolation point and $\lfloor \cdot \rfloor$ denotes the floor operator (the largest integer less than or equal to the argument). $w$ denoting the filter weight is applied by division to preserve flux:

$$w = \sum_{i=-n+1}^{n} \sum_{j=-n+1}^{n} L(i - x + \lfloor x \rfloor; n) \cdot L(j - y + \lfloor y \rfloor; n) \quad (12)$$

Lanczos interpolation utilizes the neighbourhood of the $2n \times 2n$ nearest mapped pixels. A two-dimensional Lanczos filter is non-separable. This property turns the complexity of Lanczos interpolation is $O(N \times 4n^2)$. Lancoz interpolation works on orders 3,4, and 5. In this paper, we are implementing a novel lancoz interpolation of order 5. Unlike other interpolation functions, Lanczos has infinite support with alternating positive and negative lobes around the origin.

Depending on the order, the first positive lobe is kept around the origin, the second negative lobe, the third positive lobe, and so forth. Order 4 adds the fourth negative lobe, and so on. The 2D plots in Figure 4, figure 5 and Figure 6 show the lancoz interpolation with orders 3,4 and 5.

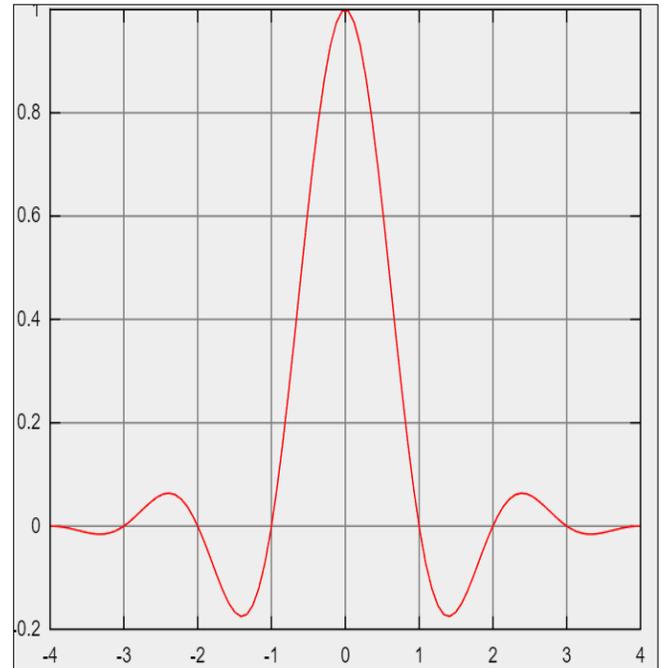

Figure 5. Lancoz interpolation of order 4.

*Harnessing The Power of Attention for Patch-Based Biomedical Image Classification*

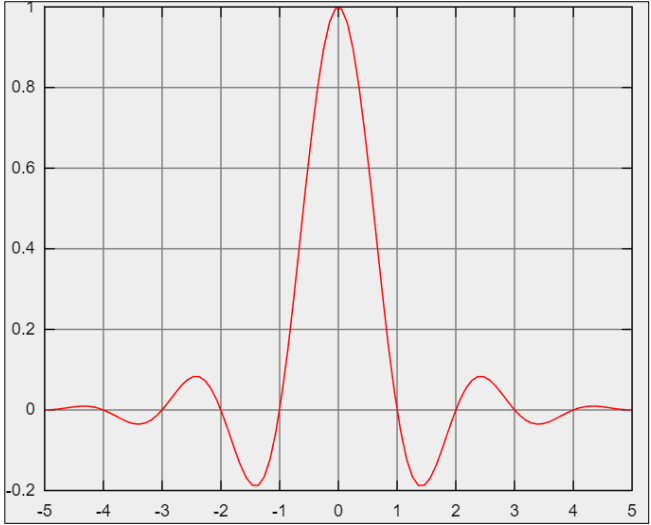

Figure 6. Lancoz interpolation of order 5.

The accuracy of lancoz3, lancoz4 and lancoz5 algorithms is 1/216 and about three times faster than their function evaluation counterparts.

## VI. RESULTS AND DISCUSSIONS

Followed by preprocessing, such as resizing and normalization. The proposed interpolation technique, Lancoz5 interpolation, is applied to the preprocessed dataset to get the interpolated images of better resolution.

After interpolation, novel cut mix data augmentation is applied. The original training dataset has been shaped as **(1260,224,224,3), (420,224,224,3), (420,224,224,3))**. The pre-augmented dataset images' snapshot is given here, as in Figure 7.

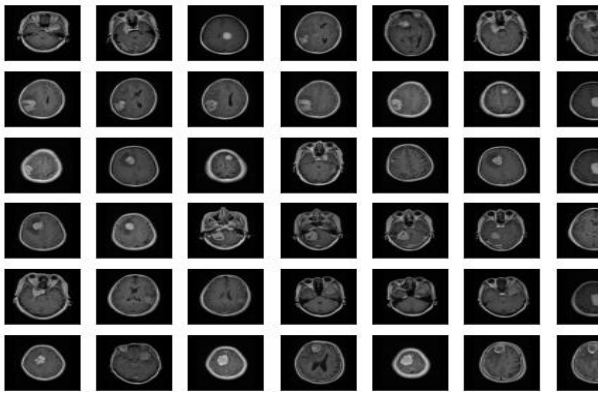

Figure 7. Snapshot of Training dataset images

After implementing the Cut Mix data augmentation technique, the augmented dataset is given in Figure 8. A brief description of the Cut Mix data augmentation technique is given:

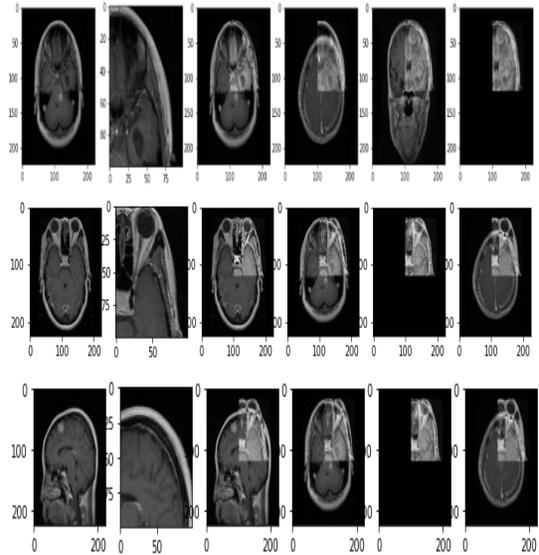

Figure 8. Augmentation using Cut Mix

After augmenting the dataset, vanilla patching is applied to the processed images to generate the visual tokens for further processing through the attention mechanism. The tokens generated after implementing vanilla patching are shown in Figure 9. The number of patches obtained is calculated using the formula:

$$\text{num\_patches} = \left(\frac{\text{image\_size}}{\text{patch\_size}}\right)^2 \quad (13)$$

**i)** *Vanilla Patching:*

After applying the vanilla patching with variable patch sizes and several patches, the patches formed are shown in the figures below.

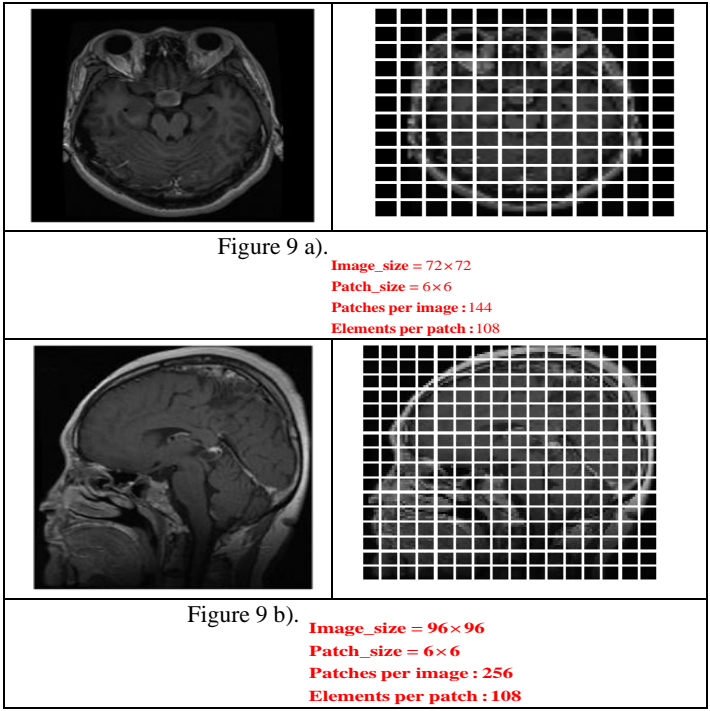

Figure 9 a).
**Image_size = 72×72**
**Patch_size = 6×6**
**Patches per image : 144**
**Elements per patch : 108**

Figure 9 b).
**Image_size = 96×96**
**Patch_size = 6×6**
**Patches per image : 256**
**Elements per patch : 108**

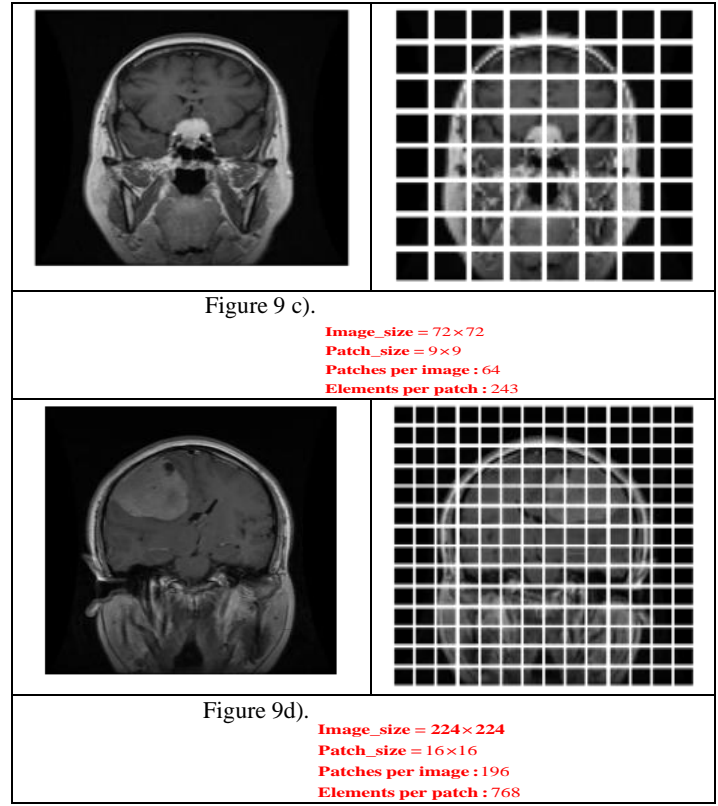

Figure 9 c).
**Image_size = 72×72**
**Patch_size = 9×9**
**Patches per image : 64**
**Elements per patch : 243**

Figure 9d).
**Image_size = 224×224**
**Patch_size = 16×16**
**Patches per image : 196**
**Elements per patch : 768**

ii). *S.P.T. Patching*

With S.P.T., an input image is shifted in several directions and concatenated with the input image. Figures 10a, 10b, 10c, and 10d show how S.P.T. patches are generated with a diagonal shift in four directions. The next step is to apply patch partitioning in the same way as standard ViTs. A three-step process is then performed to embed the data into visual tokens: patch flattening, layer normalization and linear projection. Through S.P.T., more spatial information can be embedded into visual tokens, which increases the locality inductive bias of V.I.T.s.

Vision Transformers (ViTs) with shifted patching have a much stronger inductive bias than grid-based ones. In contrast with regular grid-based patches, shifted patching involves extracting patches from a grid that has been spatially shifted over the image. In this way, we introduce translation equivariance into the model, allowing us to detect the same object or feature even if it shifts slightly.

VII. PROPOSED MODEL CONFIGURATION The model configuration is listed in Table **1.**

Table 1. Model Configuration

| Hyperparameter | Value |
|---|---|
| Learning-rate | 0.001 |
| Weight Decay | 0.0001 |
| Project Dimension | 64 |
| Number of head | 4 |
| M.L.P. Head Units | [2048,1024] |
| Number of Transformer layers | 8 |
| Layer-Norm _Eps | 1e–6 |

The number of heads and transformer layers can increase depending on available resources. Due to the limitations of the resources, we have experimented with the given environment.

VIII. TRAINING REGIME AND PERFORMANCE ANALYSIS

To effectively train ViTs, DeiT recommends the use of a variety of techniques. We, therefore, applied data augmentations such as Cut Mix, regularization technique as dropout and novel preprocessing technique as Lancoz-5 interpolation for adopting larger image sizes with better resolution. AdamW served as an optimizer while these processes were taking place. There was a 0.00001 weight decay, a batch size of 256, and a warm-up of 5. The model is trained for 100 epochs with variable patch sizes at different instances. The learning graphs (accuracy and loss curves) and R.O.C. and R.O.C. characteristic curve for each class is given in Table 2. Table 3, Table 4 and Table 5 are below. Also, the performance of the proposed model is validated using R.O.C. and R.O.C. characteristic curves shown in the Figures below from 10(a-d).

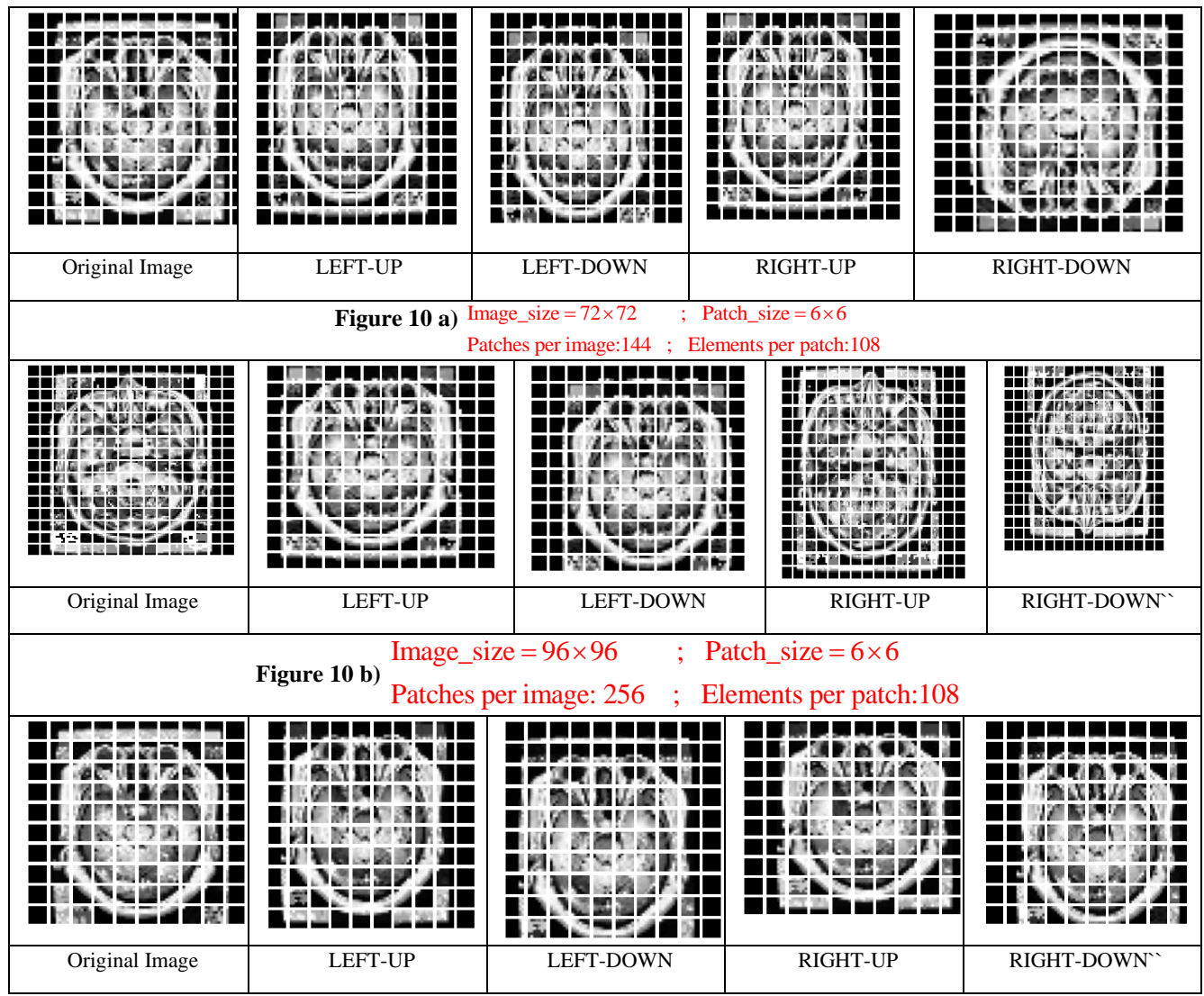

**Figure 10 a)** Image_size = $72 \times 72$ ; Patch_size = $6 \times 6$
Patches per image:144 ; Elements per patch:108

**Figure 10 b)** Image_size = $96 \times 96$ ; Patch_size = $6 \times 6$
Patches per image: 256 ; Elements per patch:108

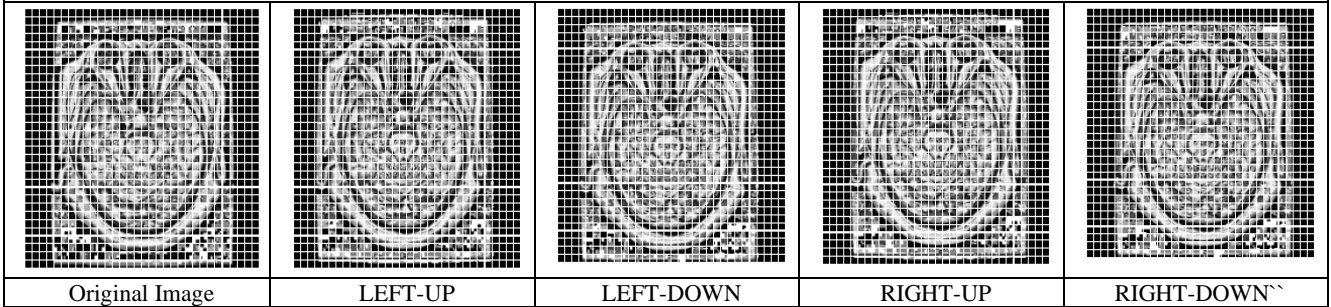

**Figure 10 c)** Image_size = 72×72 ; Patch_size = 7×7
Patches per image:100 ; Elements per patch:147

Original Image | LEFT-UP | LEFT-DOWN | RIGHT-UP | RIGHT-DOWN``

**Figure 10 d)** Image_size = 224×224 ; Patch_size = 7×7
Patches per image:1024 ; Elements per patch:147

***Case 1: 100 patches/image and 243 elements/patch***

The experiments are performed with a variable number of patches. We use 100 patches/images and 243 elements per patch in the first case. The results are given in the plots shown in Figure 11 as below:

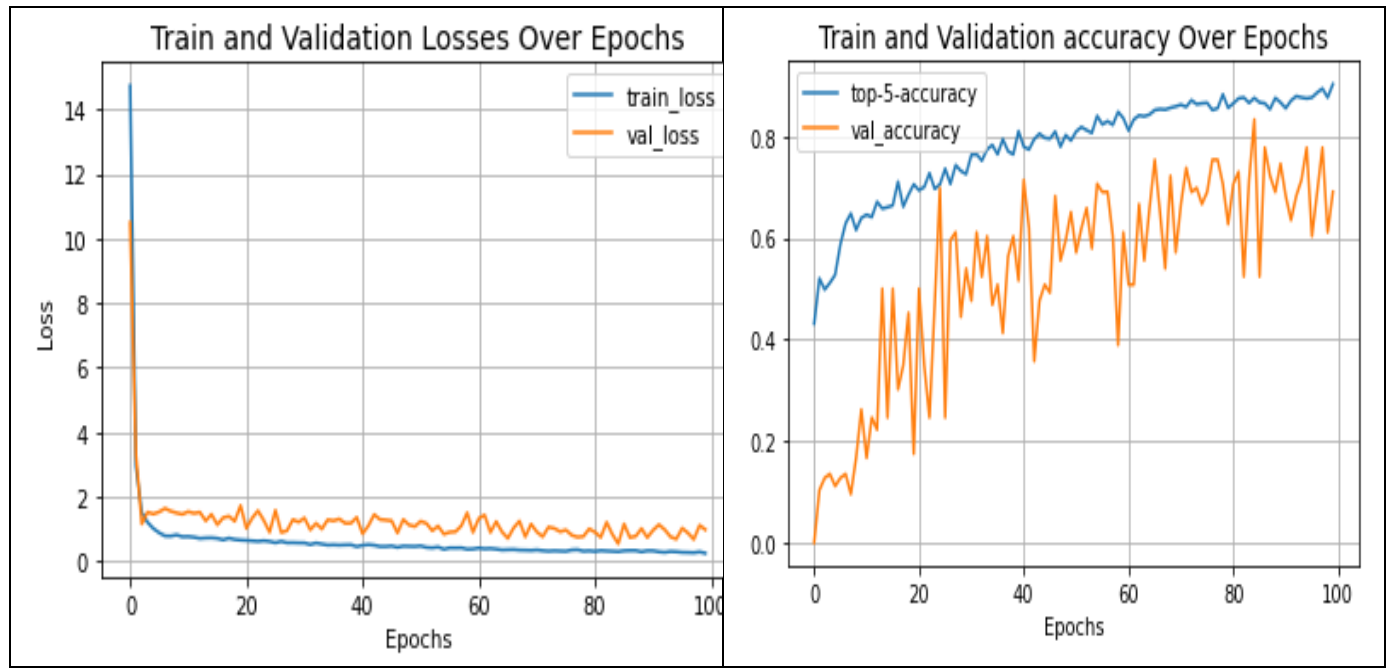

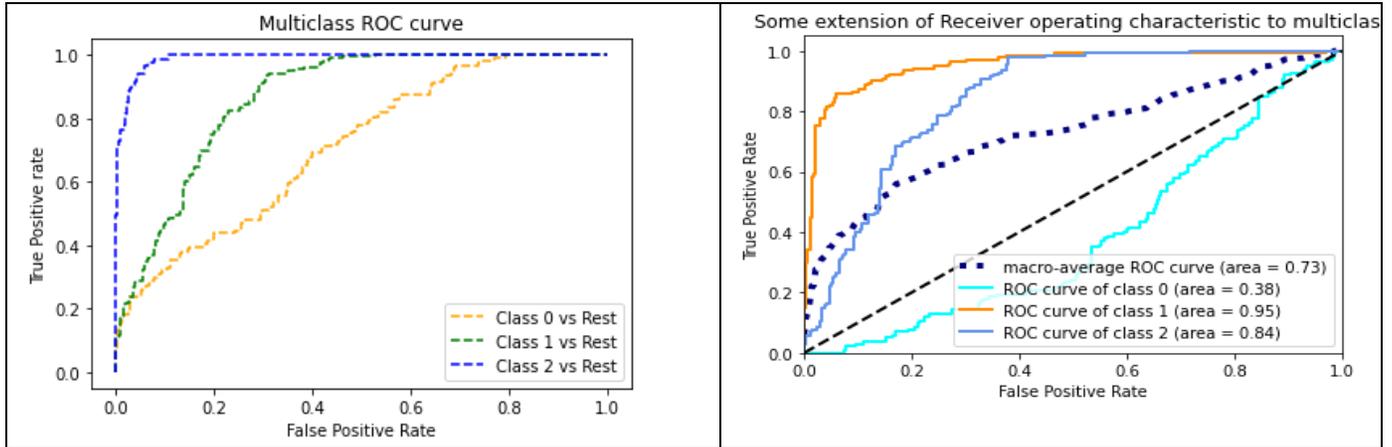

Figure 11. Learning and R.O.C. curves for case 1 144 patches/image and 108 elements/patch

*Case 2: 256 patches/image and 108 elements /patch*

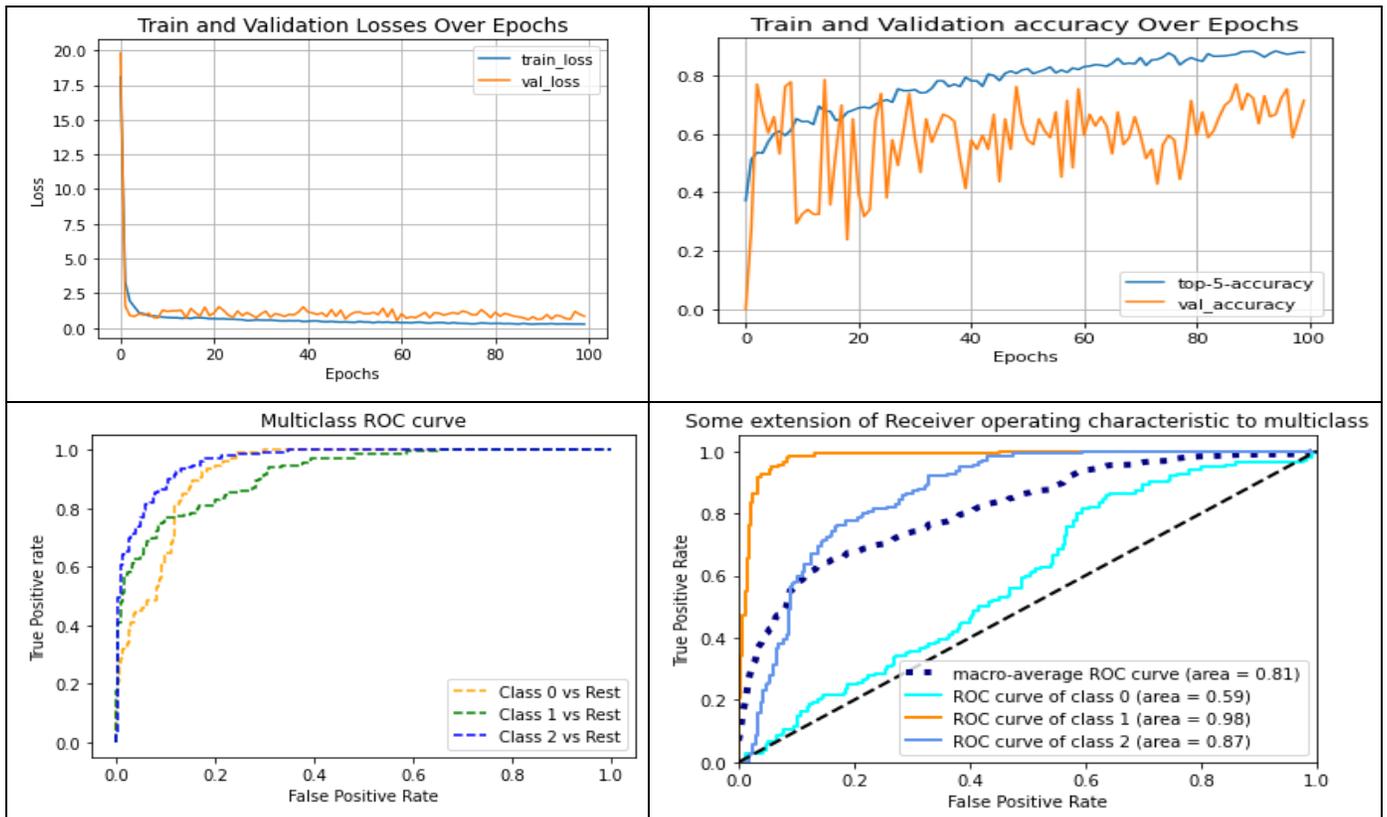

Figure 12: Learning and R.O.C. curves for case 2 256 patches/image and 108 elements /patch

*Case 3: 144 patches/image and 108 elements/patch*

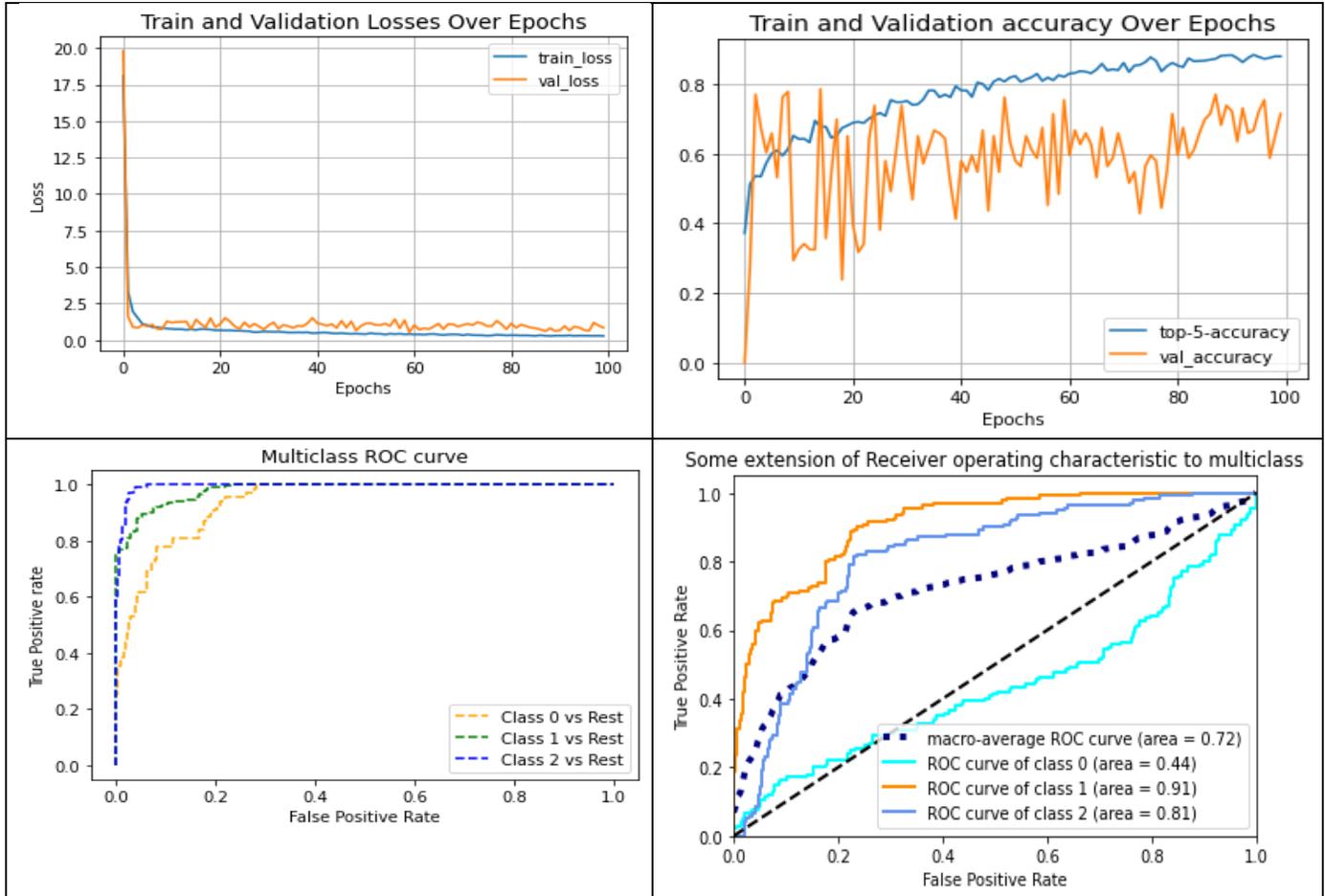

Figure 13: Learning and R.O.C. curves for case 3 64patches/image and 243 elements/patch

*Case 4:* *196patches/image and 768 elements/patch*

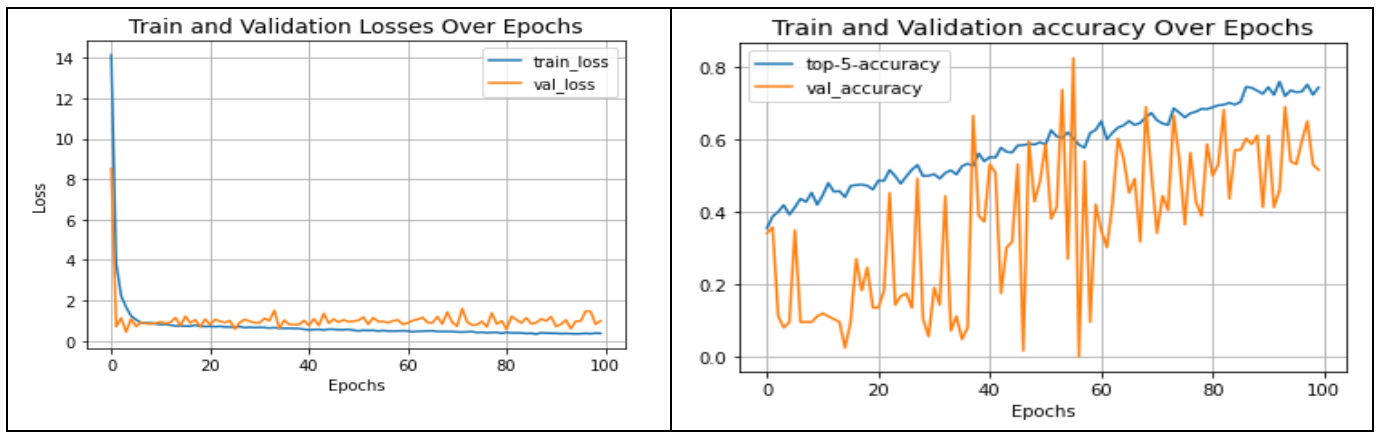

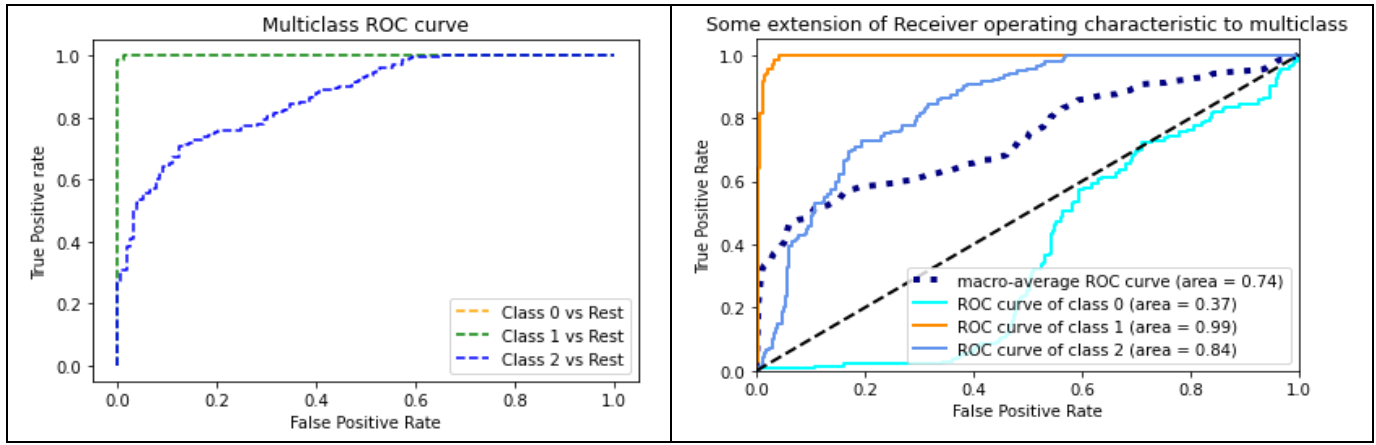

Figure 14: Learning and R.O.C. curves for case 4 196patches/image and 768 elements/patches

R.O.C. curves show trade-offs between sensitivity (or T.P.R.) and specificity (1 - F.P.R.). Better performance is demonstrated by classifiers whose curves are closer to the top-left corner. Generally, random classifiers tend to give points along the diagonal (F.P.R. = T.P.R.). As the curve approaches the diagonal of the R.O.C. space, the test becomes less accurate. The positive rate for ViT with a higher number of patches is higher than that for Vanilla patching ViT having fewer patches, as in case 1 and case 3 with 100 and 144 patches only. Throughout all cut-offs, the false positive rate is lower than that for vanilla patching with more patches. The area under the curve for ViT with a more significant number of patches is larger than that for Vanilla ViT with fewer patches. Nevertheless, vanilla patching may only partially capture fine-grained spatial information, potentially causing missing details and spatial relationships. With shifted patching, this limitation is addressed by allowing patches to overlap, retaining more context and transformer-based benefits.

.

*S.P.T. Method*

Under the same experimental setup, the S.P.T. method was trained and tested on the same dataset used for vanilla patching. The learning (loss and accuracy curves) is given in the graphs below. Also, the validation of the model is represented by the R.O.C. and R.O.C. characteristic curves presented in the plotted graphs enclosed in Table 6. For comparison, we used two patch size combinations for the S.P.T. performance analysis in Figures 15 and 16. One can go for any number of patches with different configurations to assess the performance of the proposed model.

**Case 1:** 144patches/images and 108 elements/patch

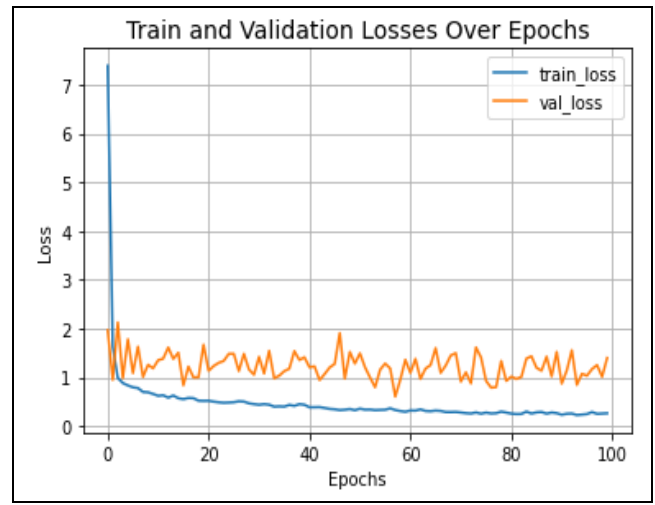

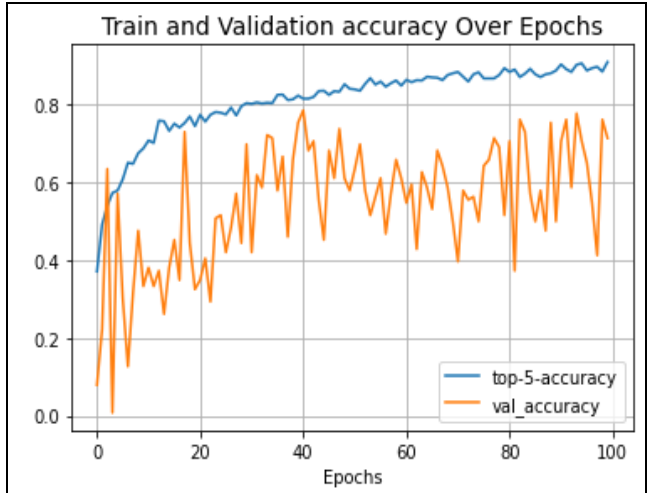

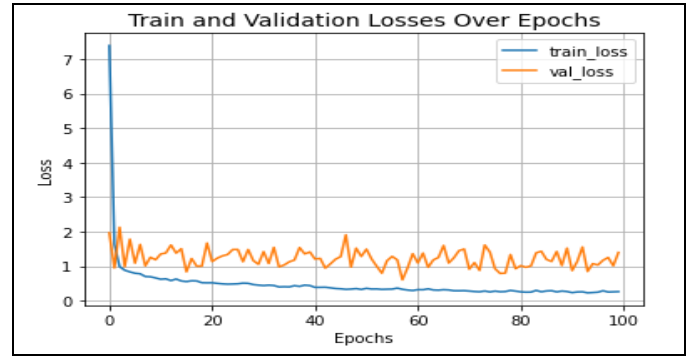

*Case 2:* *100patches/images and 147 elements/patch*

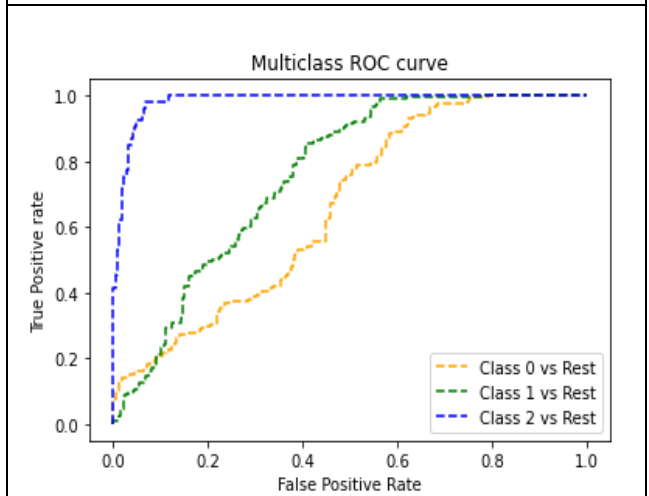

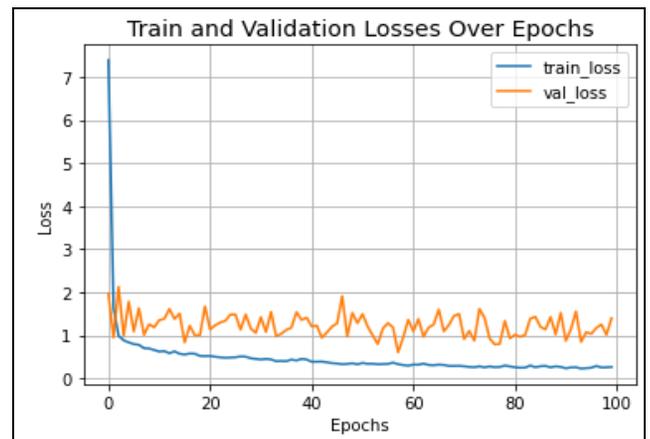

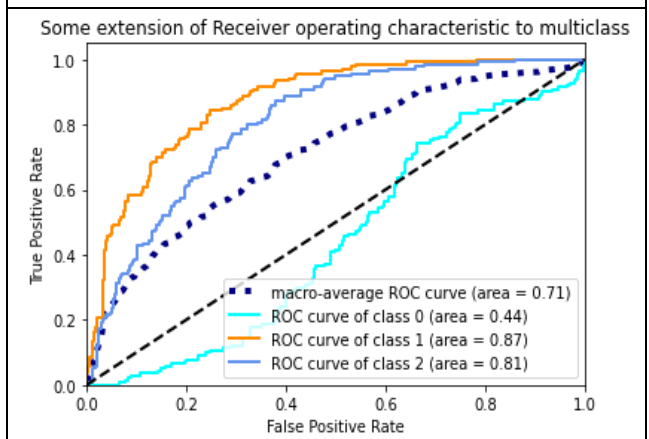

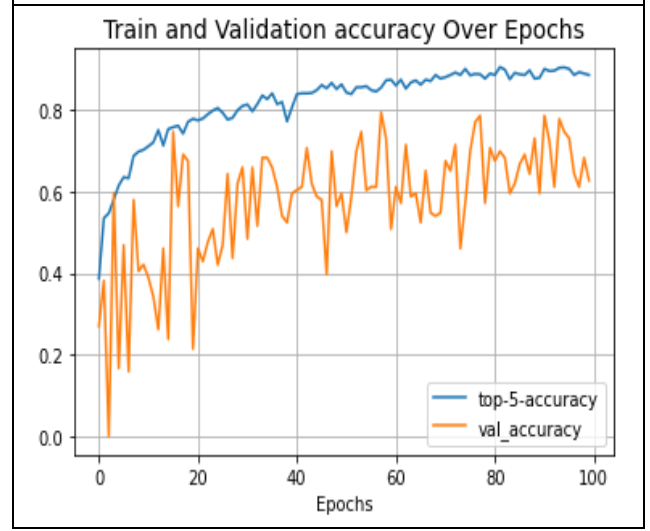

Figure 15: Case 1 Learning curves (Loss, accuracy, R.O.C., R.O.C. characteristics)

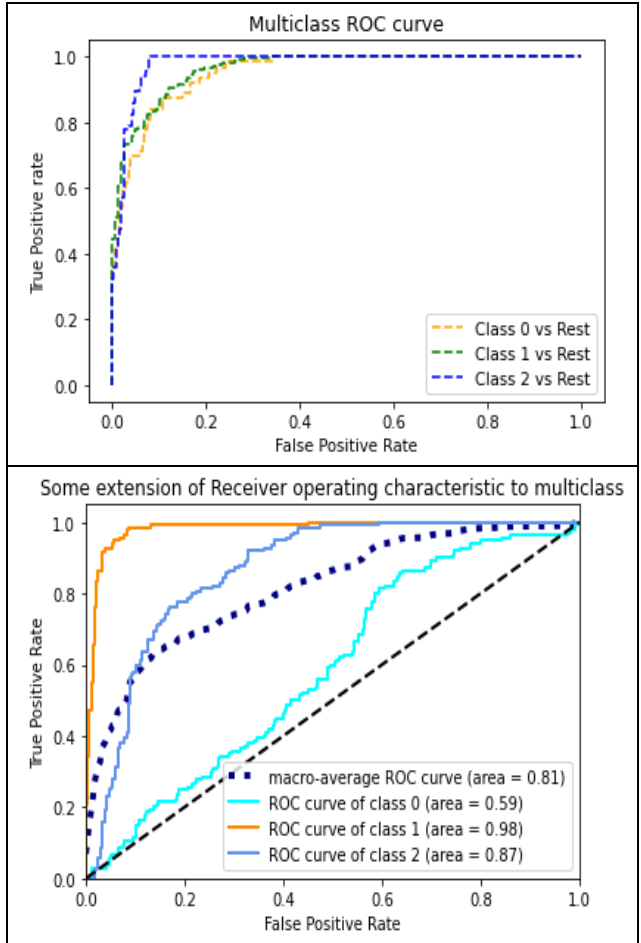

Figure 16: case 2: Learning curves (Loss, accuracy, R.O.C., R.O.C. characteristics)

***Case 3:*** *196patches/images and 768 elements/patch*

*The R.O.C. characteristics multi-class R.O.C. characteristics curves for the other two cases using S.P.T. patching technique are given below:*

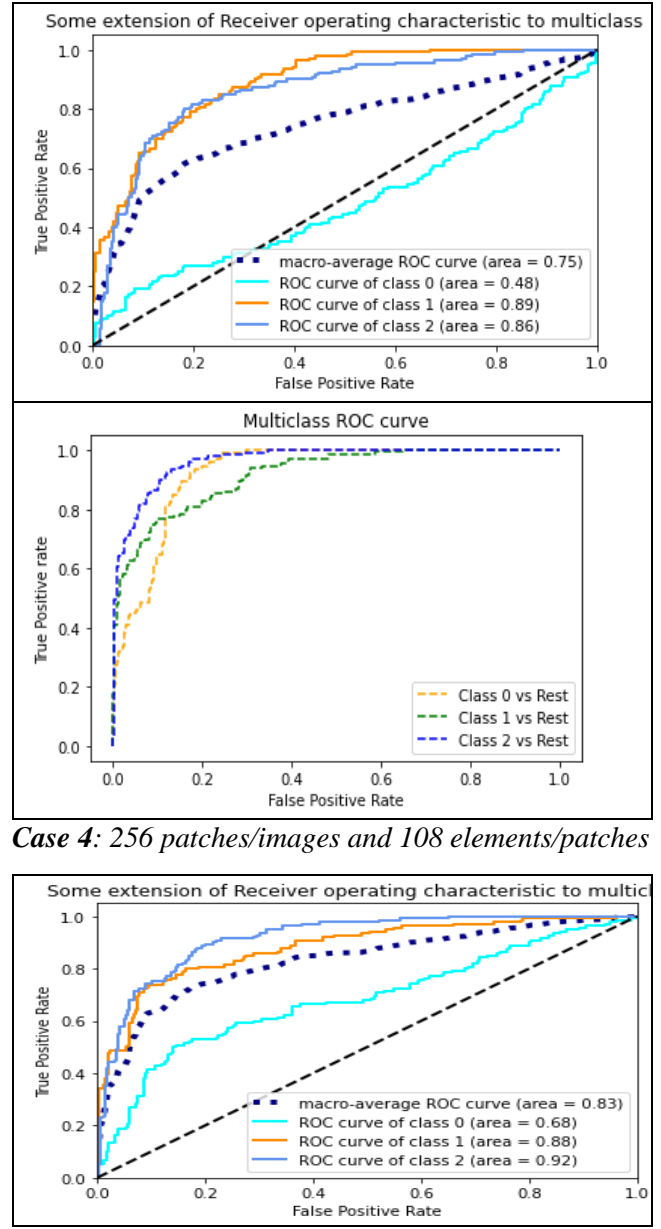

***Case 4****: 256 patches/images and 108 elements/patches*

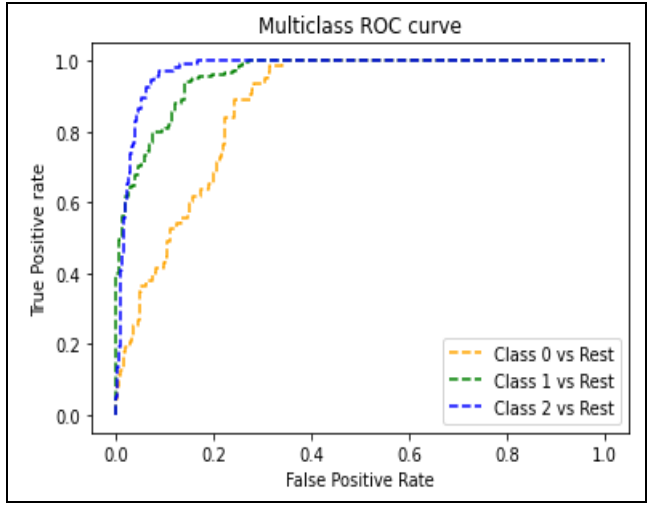

R.O.C. curves show trade-offs between sensitivity (or T.P.R.) and specificity (1 - F.P.R.). Better performance is demonstrated by classifiers whose curves are closer to the top-left corner. Generally, random classifiers tend to give points along the diagonal (F.P.R. = T.P.R.). As the curve approaches the diagonal of the R.O.C. space, the test becomes less accurate. The actual positive rate for S.P.T. ViT is higher than that for Vanilla patching ViT throughout all cut-offs, and the false positive rate is lower than that for test B. The area under the curve for S.P.T. ViT is more significant than that for Vanilla ViT. These characteristics indicate that S.P.T. ViT is outperforming Vanilla, patching ViT considerably. Also, there is increased performance gain with better test and validation accuracy for more patches. This is inferred that by increasing the number of patches locality concept of CNN is introduced in the attention mechanism to improve the performance. With enough data and a more significant number of patches, the attention mechanism behaves more like a CNN by introducing locality to each image patch and retrieving global contextual information due to its self-attention mechanism. Hence better performance gains. Having enough data can make the attention-based models learn the intrinsic properties of CNN (inductive bias and translational equivariance) from scratch. Thus, in this paper, Cut-Mix data augmentation generates aggressive augmented data.

Thus, the need for convolutions can be completely avoided. The proposed Self-attention mechanism for image classification has strong modelling capacity and, therefore, can take less time to train, even more than a few secs from scratch. This is complementary to CNN training can take days to weeks to complete training when trained from scratch. Thus, attention-based models are much computationally fast than CNN. These models are preferable to time-consuming CNN models.

IX. QUANTITIVE RESULTS

This section presents the experimental results for the small-size dataset. Test and validation accuracy of the proposed using Vanilla with the proposed Lancoz 5 interpolation method and S.P.T. patching with variable patch sizes for 100 epochs with a batch size of 256 is given in Table 8 as:

*For Vanilla patching with Lancoz 5 interpolation*

For Vanilla patching with a variable number of patches, batch size of 256, and several epochs =100. The validation and test accuracy of the model is given in Table 8:

Table 8. Vanilla patching with Lancoz 5 interpolation

| Image size | # Patches | Top-1 Test-accuracy (%) | Validation accuracy (%) | Top-5 test accuracy (%) |
|---|---|---|---|---|
| 72×72 | 144 | 66.24 | 90.77 | 91.55 |
| 72×72 | 100 | 62.48 | 78.24 | 92 |
| 224×224 | 196 | 78.48 | 96.87 | 92.64 |
| 224×224 | 256 | 94.86 | 97.64 | 92.99 |

We can conclude from the above Table that vanilla patching with the Lancoz5 interpolation method performs well, with a greater test accuracy of 94.8% and a patch size of 16*16. when using S.P.T. patching. The test and validation accuracy of the S.T.P. under the same constraints is given in Table 9:

*For S.P.T. patching*

Table.9 S.T.P. patching

| Image size | #Patches | Top-1 Test-accuracy (%) | Validation accuracy (%) | Top-5 test accuracy (%) |
|---|---|---|---|---|
| 72×72 | 144 | 63.67 | 78.04 | 93.76 |
| 72×72 | 100 | 62.57 | 64.23 | 94.12 |
| 224×224 | 196 | 79.54 | 67.44 | 95.55 |
| 224×224 | 256 | 93.87 | 96.89 | 96.36 |

In Comparing Table.8 and Table 9, the performance of the proposed model using vanilla patching is comparable to S.T.P., lagging 0.99% accuracy from the S.P.T. having a test accuracy of 93.87 with a 16*16 patch size. The performance of S.P.T. is slightly better than vanilla patching. The top-5 test accuracy is 100% in both cases outperforming the state of art models' top-5 accuracy.

X. COMPARISON WITH STATE-OF-ART

In comparing with the state of art techniques for image classification task of biomedical images (BARIN MRI) the comparative results are given in Figure 11:

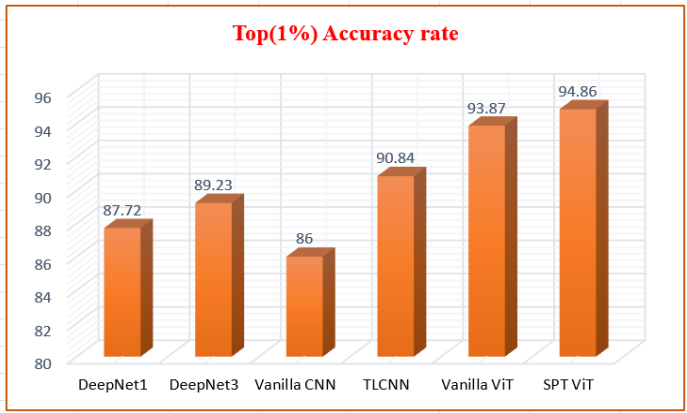

Figure 11. Comparison with state of the art

It is observed that setting the temperature factor $\sqrt{d_k}$ $\frac{1}{2}\sqrt{d_k}$ yields better results $4\sqrt{d_k}$ than, $\frac{1}{4}\sqrt{d_k}$ and after varying the temperature to $2\sqrt{d_k}$ factor, the results are slighter better than $4\sqrt{d_k}$ $\frac{1}{4}\sqrt{d_k}$. The temperature factor affects the top 1% accuracy of the model, as shown in Table 10. This can help the model avoid overfitting the training data and generalize to the test data. A high temperature encourages the model to assign similar probabilities to all classes, resulting in more diverse predictions. Temperature scaling can also improve the calibration of the model's confidence scores. By adjusting the temperature parameter, the model's predicted probabilities can be better calibrated to the actual probabilities of the classes. The temperature significantly impacts deep learning models by affecting their generalization, exploration vs. exploitation, diversity, and calibration. The choice of temperature should be carefully considered based on the specific application and the dataset's properties.

Also, the results are given in Table 11 upon testing bicubic and bilinear interpolation.

Table 11. Interpolation Techniques Comparison

From Figure 11, it is clear that the accuracy of the proposed model is comparable to the state-of-the-art, and the top 1% accuracy of both vanillas and S.P.T. ViT is outperforming the state-of-the-art.

XI. ABLATION STUDY (Effect of Temperature on Accuracy)

On varying the temperature factor, the top 1% accuracy of the model is varied, as given in Table 10.

Table 10. Ablation study on varying Temp. Factor.

| Temperature | ViT (Top-1 Test accuracy) | S.P.T. ViT (Top-1 Test accuracy) |
|---|---|---|
| $\frac{1}{4}\sqrt{d_k}$ | 75.2 | 68.32 |
| $\frac{1}{2}\sqrt{d_k}$ | 83.02 | 91.52 |
| $\sqrt{d_k}$ | 78.02 | 90.23 |
| $2\sqrt{d_k}$ | 70.34 | 88.36 |
| $4\sqrt{d_k}$ | 69.52 | 87.23 |

| Interpolation | Top-1 Test accuracy | Time complexity |
|---|---|---|
| Lancoz-5 | 84 | $O(n*4n^2)$ |
| Bicubic | 80 | $O(N*n)$ |
| Bilinear | 78 | $O(n^2)$ |

It is clear from the above Table that Lancoz5 is performing well as compared to Bicubic and Bilinear interpolation techniques. Bicubic and Bilinear interpolation techniques are discussed in detail.

The other performance metric of the proposed model is shown in Table 12.

Table 12. Throughput and Flops of the proposed model

| Methodology | Throughput(images/sec) | #Params(millions) | FLOPS (millions) |
|---|---|---|---|
| ViT | 8593 | 2.8M | 189.8 |
| SPT ViT | 6632 | 8.7M | 280.4 |

Table 11 shows that the throughput of the ViT model is relatively high compared to the ViT model with S.P.T.

XII. HYPOTHESIS AND LIMITATIONS OF THE PROPOSED METHOD

This work uses attention-based models to improve model capacity and generalization. We tried to completely replace the state-of-art CNN models with attention-based models, but these models also face some critical challenges that need to get addressed.

There is often a need for a large amount of training data or extra supervision in pure attention-based model architectures to achieve comparable performance with CNNs.

XIII. FUTURE DIRECTIONS

- To overcome the drawbacks, we must develop a solution that combines the CNN intrinsic features (locality, weight sharing, inductive biases) to improve the overall performance of these attention-based models.

- Despite the attention-based visual model's powerful ability to model global features, CNN can effectively process low-level features to improve the locality of the visual Transformers by padding and adding positional features via padding.

- The model needs to be more parameterized, and the number of FLOPS is high. We must look for an effective and aggressive model compression technique to reduce the number of parameters and FLOPS without compromising the model's actual performance.

- We can induce the intrinsic properties of the convolution into attention-based vision models via distillation to improve the inductive bias of the models. This leads to an increased generalization of the model.

Convolutional networks, combined with self-attention, enable a comprehensive approach to visual understanding that encompasses local and global patterns, making it ideal for tasks involving complex visual data, such as videos and images.

XIV. CONCLUSION

CNNs are primarily limited by spatial dependencies when they are used in the context of applied research. The paper proposes a more effective solution to this problem by adopting self-attention mechanisms for vision tasks. We present a comprehensive analysis of the self-attention mechanism, emphasizing its potential to capture long-range dependencies. Three innovative techniques are introduced in the study: Cut-Mix, Lancoz5, and S.P.T. tokenization, all designed to enhance the inductive bias of attention-based models. Extensive experiments demonstrate that increasing the number of patches enhances the localized context of images, thus enhancing performance. Copious data is generated by incorporating Cut-Mix, enabling attention-based models to learn CNN-like intrinsic properties. Ultimately, the research establishes a compelling proposition: the feasibility of biomedical image classification without conventional convolution. In addition to exhibiting exceptional prowess in analogous tasks, attention-based vision models have the powerful modelling capacity to replace convolutions.

As a result of this research, CNNs are being replaced by attention-based models, augmented by innovative techniques and rigorous experiments, which transcend the limitations of CNNs. This way, convolutions can be dispensable while maintaining performance and creating a more comprehensive understanding of visual data. Below is the flow chart illustrating the entire methodology used.

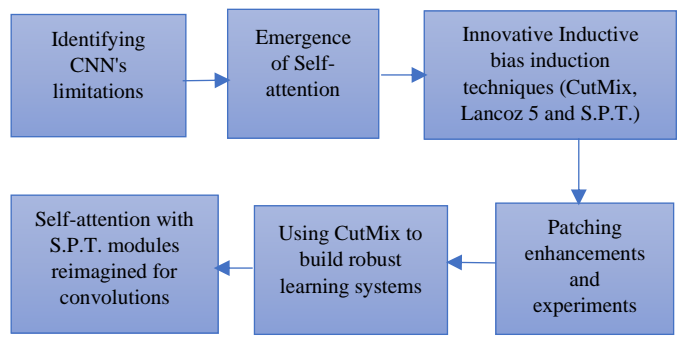

Figure 12: Flowchart of proposed Methodology

**Author Contributions:** Gousia Habib & Shaima Qureshi: Conceptualization and design, formal analysis, Coding and data analysis, writing and editing.

**Data availability:** Data is available from

https://figshare.com/articles/dataset/brain_tumor_dataset/1512427.

**Funding:** Not applicable.

**Code availability:** (will be available soon)

**Declarations**

**Conflicts of Interests:** The authors declare that they have no conflicts of interest or competing interests relating to the content of this article.

**Ethics approval**: This article contains no studies with human participants or animals performed by authors.

**Consent to participate:** Not applicable. Consent for publication is Not applicable.

REFERENCES

[1]. Sellami, A., & Tabbone, S. (2022). Deep neural networks-based relevant latent representation learning for hyperspectral image classification. *Pattern Recognition*, 121, 108224. https://doi.org/10.1016/j.patcog.2021.108224

[2]. Zhong, H., Lv, Y., Yuan, R., & Yang, D. (2022). Bearing fault diagnosis using transfer learning and self-attention ensemble lightweight convolutional neural network. *Neurocomputing*, *501*, 765-777. https://doi.org/10.1016/j.neucom.2022.06.066

[3]. Tunstall, L., Von Werra, L., & Wolf, T. (2022). Natural Language Processing with Transformers, Revised Edition. O'Reilly Media, Incorporated.

[4]. Majid, S., Alenezi, F., Masood, S., Ahmad, M., Gündüz, E. S., & Polat, K. (2022). Attention-based CNN model for fire detection and localization in real-world images. Expert Systems with Applications, 189, 116114. https://doi.org/10.1016/j.eswa.2021.116114

[5]. Wang, Y., Chen, Z., & Chen, Z. B. (2022). Dynamic graph Conv-LSTM model with dynamic positional encoding for the large-scale travelling salesman problem. Mathematical Biosciences and Engineering, 19(10), 9730-9748. https://doi.org/10.3934/mbe.2022452

[6] Krizhevsky, A., & Hinton, G. (2009). Learning multiple layers of features from tiny images.

[7] Revisiting the unreasonable effectiveness of data. https://ai.googleblog. com/2017/07/revisiting-unreasonable-effectiveness.html.

[8]. Deng, J., Dong, W., Socher, R., Li, L. J., Li, K., & Fei-Fei, L. (2009, June). Imagenet: A large-scale hierarchical image database. In 2009 IEEE conference on computer vision and pattern recognition (pp. 248-255). Ieee. https://doi.org/10.1109/CVPR.2009.5206848

[9]. Dong, B., Zeng, F., Wang, T., Zhang, X., & Wei, Y. (2021). Souq: Segmenting objects by learning queries. Advances in Neural Information Processing Systems, 34, 21898-21909. https://doi.org/10.48550/arXiv.2106.02351.

[10]. Gulati, A., Qin, J., Chiu, C. C., Parmar, N., Zhang, Y., Yu, J., ... & Pang, R. (2020). Conformer: Convolution-augmented Transformer for speech recognition. arXiv

preprint arXiv:2005.08100. https://doi.org/10.48550/arXiv.2005.08100.

[11]. Chen, Y., Fan, H., Xu, B., Yan, Z., Kalantidis, Y., Rohrbach, M., ... & Feng, J. (2019). Drop an octave: Reducing spatial redundancy in convolutional neural networks with octave convolution. In Proceedings of the IEEE/CVF International Conference on Computer Vision (pp. 3435-3444). https://doi.org/10.48550/arXiv.1904.05049.

[12]. Chen, C. F., Fan, Q., Mallinar, N., Sercu, T., & Feris, R. (2018). Big-little net: An efficient multiscale feature representation for visual and speech recognition. arXiv preprint arXiv:1807.03848. https://doi.org/10.48550/arXiv.1807.03848.

[12]. He, K., Zhang, X., Ren, S., & Sun, J. (2016, October). Identity mappings in deep residual networks. In European conference on computer vision (pp. 630-645). Springer, Cham. https://doi.org/10.1007/978-3-319-46493-0_38

[13]. Krizhevsky, A., Sutskever, I., & Hinton, G. E. (2017). Imagenet classification with deep convolutional neural networks. Communications of the A.C.M., 60(6), 84-90. https://doi.org/10.1145/3065386

[14]. LeCun, Y., Boser, B., Denker, J. S., Henderson, D., Howard, R. E., Hubbard, W., & Jackel, L. D. (1989). Backpropagation applied to handwritten zip code recognition. Neural Computation, 1(4), 541-551. https://doi.org/10.1162/neco.1989.1.4.541

[15]. Radosavovic, I., Kosaraju, R. P., Girshick, R., He, K., & Dollár, P. (2020). Designing network design spaces. In Proceedings of the IEEE/CVF conference on computer vision and pattern recognition (pp. 10428-10436).

https://doi.org/10.48550/arXiv.2003.13678

[16]. Simonyan, K., & Zisserman, A. (2014). Very deep convolutional networks for large-scale image recognition. arXiv preprint arXiv:1409.1556.

https://doi.org/10.48550/arXiv.1409.1556

[17]. Han, S., Pool, J., Tran, J., & Dally, W. (2015). Learning both weights and connections for the efficient neural network. Advances in neural information processing systems, 28. https://doi.org/10.48550/arXiv.1506.02626

[18]. Gao, S. H., Cheng, M. M., Zhao, K., Zhang, X. Y., Yang, M. H., & Torr, P. (2019). Res2net: A new multiscale backbone architecture. IEEE transactions on pattern analysis and machine intelligence, 43(2), 652-662. https://doi.org/10.1109/TPAMI.2019.2938758

[19]. Zhang, H., Wu, C., Zhang, Z., Zhu, Y., Lin, H., Zhang, Z., ... & Smola, A. (2022). Reset Split-attention networks. In Proceedings of the IEEE/CVF Conference on Computer Vision and Pattern Recognition (pp. 2736-2746).

https://doi.org/10.48550/arXiv.2004.08955

[20]. Simonyan, K., & Zisserman, A. (2014). Two-stream convolutional networks for action recognition in videos. *Advances in neural information processing systems*, *27*. https://doi.org/10.48550/arXiv.1406.2199

[21]. Feichtenhofer, C., Pinz, A., & Zisserman, A. (2016). Convolutional two-stream network fusion for video action recognition. In *Proceedings of the IEEE conference on computer vision and pattern recognition* (pp. 1933-1941).

https://doi.org/10.48550/arXiv.1604.06573

[22]. Carreira, J., & Zisserman, A. (2017). Quo Vadis, action recognition? A new model and the kinetics dataset. In *Proceedings of the IEEE Conference on Computer Vision and Pattern Recognition* (pp. 6299-6308). https://doi.org/10.1109/CVPR.2017.502

[23]. Qiu, Z., Yao, T., & Mei, T. (2017). Learning Spatiotemporal representation with pseudo-3d residual networks. In *Proceedings of the IEEE International Conference on Computer Vision* (pp. 5533-5541).

https://doi.org/10.48550/arXiv.1711.10305

[24]. Li, Z., Gavrilyuk, K., Gavves, E., Jain, M., & Snoek, C. G. (2018). Videos convolve, attend and flow for action recognition. *Computer Vision and Image Understanding*, *166*, 41-50. https://doi.org/10.1016/j.cviu.2017.10.011

[25]. Xie, S., Sun, C., Huang, J., Tu, Z., & Murphy, K. (2018). Rethinking spatiotemporal feature learning: Speed-accuracy trade-offs in video classification. In *Proceedings of*

the European Conference on computer vision (E.C.C.V.) (pp. 305-321). https://doi.org/10.48550/arXiv.1712.04851

[26]. Tran, D., Wang, H., Torresani, L., & Feiszli, M. (2019). Video classification with channel-separated convolutional networks. In *Proceedings of the IEEE/CVF International Conference on Computer Vision* (pp. 5552-5561).

https://doi.org/10.48550/arXiv.1904.02811

[27]. Feichtenhofer, C., Fan, H., Malik, J., & He, K. (2019). Slow, fast networks for video recognition. In *Proceedings of the IEEE/CVF international conference on computer vision* (pp. 6202-6211). https://doi.org/10.1109/ICCV.2019.00630

[28]. Wu, C. Y., Feichtenhofer, C., Fan, H., He, K., Krahenbuhl, P., & Girshick, R. (2019). Long-term feature banks for detailed video understanding. In *Proceedings of the IEEE/CVF Conference on Computer Vision and Pattern Recognition* (pp. 284-293). https://doi.org/10.1109/CVPR.2019.00037

[29]. Girdhar, R., Carreira, J., Doersch, C., & Zisserman, A. (2019). Video action transformer network. In *Proceedings of the IEEE/CVF conference on computer vision and pattern recognition* (pp. 244-253).

https://doi.org/10.48550/arXiv.1812.02707

[30]. Feichtenhofer, C. (2020). X3d: Expanding architectures for efficient video recognition. In *Proceedings of the IEEE/CVF Conference on Computer Vision and Pattern Recognition* (pp. 203-213). https://doi.org/10.1109/CVPR42600.2020.00028

[31]. Zhou, B., Andonian, A., Oliva, A., & Torralba, A. (2018). Temporal relational reasoning in videos. In *Proceedings of the European Conference on computer vision (E.C.C.V.)* (pp. 803-818). https://doi.org/10.1007/978-3-030-01246-5_49

[32]. Jiang, B., Wang, M., Gan, W., Wu, W., & Yan, J. (2019). Stm: Spatiotemporal and motion encoding for action recognition. In *Proceedings of the IEEE/CVF International Conference on Computer Vision* (pp. 2000-2009). https://doi.org/10.1109/ICCV.2019.00209

[33]. Ramachandran, P., Parmar, N., Vaswani, A., Bello, I., Levskaya, A., & Shlens, J. (2019). Standalone self-attention in vision models. *Advances in Neural Information Processing Systems*, *32*. https://doi.org/10.48550/arXiv.1906.05909

[34]. Zhao, H., Jia, J., & Koltun, V. (2020). Exploring self-attention for image recognition. In *Proceedings of the IEEE/CVF Conference on Computer Vision and Pattern Recognition* (pp. 10076-10085). https://doi.org/10.1109/CVPR42600.2020.01009

[35]. Hu, H., Zhang, Z., Xie, Z., & Lin, S. (2019). Local relation networks for image recognition. In *Proceedings of the IEEE/CVF International Conference on Computer Vision* (pp. 3464-3473). https://doi.org/10.1109/ICCV.2019.00356

[36]. Lu, J., Batra, D., Parikh, D., & Lee, S. (2019). Vilbert: Pretraining task-agnostic Visio linguistic representations for vision-and-language tasks. *Advances in neural information processing systems*, *32*.

https://doi.org/10.48550/arXiv.1908.02265

[37]. Li, L. H., Yatskar, M., Yin, D., Hsieh, C. J., & Chang, K. W. (2019). Visualbert: A simple and performant baseline for vision and language. *arXiv preprint arXiv:1908.03557*.

https://doi.org/10.48550/arXiv.1908.03557

[38]. Bello, I., Zoph, B., Vaswani, A., Shlens, J., & Le, Q. V. (2019). Attention augmented convolutional networks. In *Proceedings of the IEEE/CVF international conference on computer vision* (pp. 3286-3295).

https://doi.org/10.48550/arXiv.1904.09925

[39]. Henschel, L., Kügler, D., & Reuter, M. (2022). FastSurferVINN: Building resolution-independence into deep learning segmentation methods—A solution for HighRes brain M.R.I. NeuroImage, 251, 118933. https://doi.org/10.1016/j.neuroimage.2022.118933

[40]. Liu, J., Luo, H., & Liu, H. (2022). Deep learning-based data analytics for safety in construction. Automation in Construction, 140, 104302. https://doi.org/10.1016/j.autcon.2022.104302

[41]. Xu, Y., Du, B., & Zhang, L. (2021). Self-attention context network: Addressing the threat of adversarial attacks for hyperspectral image classification. IEEE Transactions on Image Processing, 30, 8671-8685. https://doi.org/10.1109/TIP.2021.3118977

[42]. Rymarczyk, D., Borowa, A., Tabor, J., & Zielinski, B. (2021). Kernel self-attention for weakly-supervised image classification using deep multiple instance learning. In Proceedings of the IEEE/CVF Winter Conference on Applications of Computer Vision (pp. 1721-1730). https://doi.org/10.1109/WACV48630.2021.00176

[43]. Li, Z., Yuan, L., Xu, H., Cheng, R., & Wen, X. (2020, December). Deep multi-instance learning with induced self-attention for medical image classification. 2020 IEEE International Conference on Bioinformatics and Biomedicine (B.I.B.M.) (pp. 446-450). IEEE. https://doi.org/10.1109/BIBM49941.2020.9313518

[44]. Qing, Y., Huang, Q., Feng, L., Qi, Y., & Liu, W. (2022). Multiscale Feature Fusion Network Incorporating 3D Self-Attention for Hyperspectral Image Classification. Remote Sensing, 14(3), 742. https://doi.org/10.3390/rs14030742

[45]. Wang, X., Girshick, R., Gupta, A., & He, K. (2018). Non-local neural networks. In Proceedings of the IEEE conference on computer vision and pattern recognition (pp. 7794-7803). https://doi.org/10.1109/CVPR.2018.00813

[47]. Hu, H., Gu, J., Zhang, Z., Dai, J., & Wei, Y. (2018). Relation networks for object detection. In Proceedings of the IEEE conference on computer vision and pattern recognition (pp. 3588-3597).

https://doi.org/10.48550/arXiv.1711.11575

[48]. Vaswani, A., Shazeer, N., Parmar, N., Uszkoreit, J., Jones, L., Gomez, A. N., ... & Polosukhin, I. (2017). Attention is all you need. Advances in neural information processing systems, 30.

https://doi.org/10.48550/arXiv.1706.03762

[49]. Dosovitskiy, A., Beyer, L., Kolesnikov, A., Weissenborn, D., Zhai, X., Unterthiner, T., ... & Houlsby, N. (2020). An image is worth 16x16 words: Transformers for image recognition at scale. arXiv preprint arXiv:2010.11929.

https://doi.org/10.48550/arXiv.2010.11929

[50]. Carion, N., Massa, F., Synnaeve, G., Usunier, N., Kirillov, A., & Zagoruyko, S. (2020, August). End-to-end object detection with transformers. In European conference on computer vision (pp. 213-229). Springer, Cham. https://doi.org/10.1007/978-3-030-58452-8_13

[52]. Khan, S., Naseer, M., Hayat, M., Zamir, S. W., Khan, F. S., & Shah, M. (2022). Transformers in vision: A survey. A.C.M. computing surveys (C.S.U.R.), 54(10s), 1-41. https://doi.org/10.1145/3505244

[53]. Wu, H., Xiao, B., Codella, N., Liu, M., Dai, X., Yuan, L., & Zhang, L. (2021). Cvt: Introducing convolutions to vision transformers. In Proceedings of the IEEE/CVF International Conference on Computer Vision (pp. 22-31).

https://doi.org/10.48550/arXiv.2103.15808

[54]. Beal, J., Kim, E., Tzeng, E., Park, D. H., Zhai, A., & Kislyuk, D. (2020). Toward transformer-based object detection. arXiv preprint arXiv:2012.09958. https://doi.org/10.48550/arXiv.2012.09958.

[55]. Zhao, H., Jiang, L., Jia, J., Torr, P. H., & Koltun, V. (2021). Point transformer. In Proceedings of the IEEE/CVF International Conference on Computer Vision (pp. 16259-16268). https://doi.org/10.48550/arXiv.2012.09164.

[56]. Wang, H., Zhu, Y., Adam, H., Yuille, A., & Chen, L. C. (2021). Max-deep lab: End-to-end panoptic segmentation with mask transformers. In Proceedings of the IEEE/CVF conference on computer vision and pattern recognition (pp. 5463-5474). https://doi.org/10.48550/arXiv.2012.00759.

[57]. Lin, K., Wang, L., & Liu, Z. (2021). End-to-end human pose and mesh reconstruction with transformers. In Proceedings of the IEEE/CVF Conference on Computer Vision and Pattern Recognition (pp. 1954-1963). https://doi.org/10.48550/arXiv.2012.09760.

[58]. Yang, S., Quan, Z., Nie, M., & Yang, W. (2021). Transpose: Keypoint localization via Transformer. In Proceedings of the IEEE/CVF International Conference on Computer Vision (pp. 11802-11812). https://doi.org/10.48550/arXiv.2012.14214.

[59]. Chen, J., Lu, Y., Yu, Q., Luo, X., Adeli, E., Wang, Y., ... & Zhou, Y. (2021). Transnet: Transformers make strong encoders for medical image segmentation. arXiv preprint

arXiv:2102.04306. https://doi.org/10.48550/arXiv.2102.04306.

[60]. Chen, M., Radford, A., Child, R., Wu, J., Jun, H., Luan, D., & Sutskever, I. (2020, November). Generative pretraining from pixels. In International conference on machine learning (pp. 1691-1703). P.M.L.R.

[61]. Guo, M. H., Cai, J. X., Liu, Z. N., Mu, T. J., Martin, R. R., & Hu, S. M. (2021). Pct: Point cloud transformer. Computational Visual Media, 7(2), 187-199. https://doi.org/10.1007/s41095-021-0229-5.

[62]. Wang, Y., Xu, Z., Wang, X., Shen, C., Cheng, B., Shen, H., & Xia, H. (2021). End-to-end video instance segmentation with transformers. In Proceedings of the IEEE/CVF Conference on Computer Vision and Pattern Recognition (pp. 8741-8750). https://doi.org/10.48550/arXiv.2011.14503.

[63]. He, S., Luo, H., Wang, P., Wang, F., Li, H., & Jiang, W. (2021). Transreid: Transformer-based object re-identification. In Proceedings of the IEEE/CVF international conference on computer vision (pp. 15013-15022). https://doi.org/10.48550/arXiv.2102.04378.

[64]. Gabeur, V., Sun, C., Alahari, K., & Schmid, C. (2020, August). Multimodal Transformer for video retrieval. In European Conference on Computer Vision (pp. 214-229). Springer, Cham. https://doi.org/10.48550/arXiv.2007.10639.

[65]. Le, H., Sahoo, D., Chen, N. F., & Hoi, S. C. (2019). Multimodal transformer networks for end-to-end video-grounded dialogue systems. arXiv preprint arXiv:1907.01166. https://doi.org/10.18653/v1/P19-1564.

[66]. Yuan, Z., Song, X., Bai, L., Wang, Z., & Ouyang, W. (2021). Temporal-channel Transformer for 3d lidar-based video object detection for autonomous driving. IEEE Transactions on Circuits and Systems for Video Technology, 32(4), 2068-2078. https://doi.org/10.48550/arXiv.2011.13628.

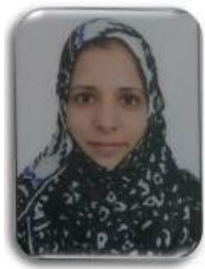

**Gousia Habib** Currently a Postdoctoral Research fellow at Indian Institute of Technology Delhi(IIT Delhi) .Perused her PhD in computer science and engineering from the National Institute of Technology Srinagar. M.Tech from the Central University of Punjab in 2018. B. Tech from Kashmir University in 2015.

Areas of research include Computer vision, machine learning, optimization, deep learning, Biomedical image analysis, pattern recognition and N.L.P., Computer Vision and Generative AI.

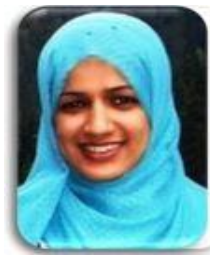

**Shaima Qureshi** received a Doctor of Philosophy (PhD) from the National Institute of Technology Srinagar (N.I.T. Srinagar. She has completed her B.E. B.E. (Hons.) Computer Science degree from BITS Pilani, India, in 2004.

She completed her M.S.M.S. in Computer Science from Syracuse University, NY, U.S.A., in 2006. She has been working as an Assistant Professor at N.I.T. Srinagar since 2008. She has been guiding PhD and B.E.B.E. students and has several patents and publications to her credit. Before joining the academic field, she worked as a Senior QA Engineer for two years in the software industry in the U.S.A. Her research areas include Mobile Networks, Algorithms, Operating Systems and Machine Learning.

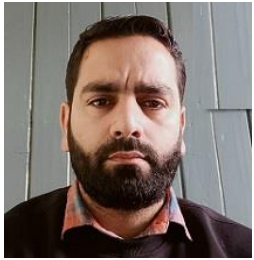 **Ishfaq Ahmad Malik** perused his PhD in Mathematics from the National Institute of Technology Srinagar. M.Sc. Mathematics from University of Kashmir. Qualified NET with AIR-013 in 2014.

Areas of research include Summability, Differential Equations, and Data Analysis.